%% file: main.tex
\documentclass[review,longtitle, 3p]{elsarticle}

\usepackage{hyperref}

\journal{Medical Image Analysis}

\usepackage{hyperref}
\usepackage{eqnarray}
\usepackage{multirow}
\usepackage{pifont}
\usepackage{todonotes}
\usepackage{lscape}
\usepackage{subfig}
\usepackage{amsmath}

\newcommand{\cmark}{\ding{51}}%
\newcommand{\xmark}{\ding{55}}%

\newcolumntype{L}[1]{>{\raggedright\let\newline\\\arraybackslash\hspace{0pt}}m{#1}}
\newcolumntype{C}[1]{>{\centering\let\newline\\\arraybackslash\hspace{0pt}}m{#1}}
\newcolumntype{R}[1]{>{\raggedleft\let\newline\\\arraybackslash\hspace{0pt}}m{#1}}

\biboptions{authoryear}

\begin{document}

\begin{frontmatter}

\title{REFUGE~Challenge: A Unified Framework for Evaluating Automated~Methods for Glaucoma~Assessment from Fundus Photographs}


\input{authors.tex}

\input{abstract.tex}


\end{frontmatter}

\input{abbreviations.tex}
\input{introduction.tex}

\input{previous_works.tex}

\input{evaluation_framework.tex}

\input{results.tex}

\input{discussion.tex}

\input{conclusions.tex}

\section*{Acknowledgements}

This work was supported by the Christian Doppler Research Association, the Austrian Federal Ministry for Digital and Economic Affairs and the National Foundation for Research, Technology and Development, J.I.O is supported by WWTF (Medical University of Vienna: AugUniWien/FA7464A0249, University of Vienna: VRG12-009). Team Masker is supported by Natural Science Foundation of Guangdong Province of China (Grant 2017A030310647). Team BUCT is partially supported by the National Natural Science Foundation of China (Grant 11571031). The authors would also like to thank REFUGE study group for collaborating with this challenge. 

\section*{Conflicts of Interest}

None.


\bibliography{references}

\appendix
\section{Participating methods}

\subsection*{\textbf{AIML}}

\paragraph{\textbf{OD/OC segmentation}} The method was a two-stage approach based on a combination of multiple dilated fully-convolutional networks (FCNs) based on ResNet-50, -101, -152~\citep{he2016deep} and -38~\citep{wu2019wider}. First, a ResNet-50 FCN was used to coarsely segment the ONH. The corresponding region was afterwards cropped to cover approximately one quarter of the original resolution. These images were used to fed the ResNet-50, -101, -152~\citep{he2016deep} and -38~\citep{wu2019wider} models, which produced the final segmentations of the OD/OC. The networks were trained using the REFUGE training set with data augmentation, including rescalings and rotations. The final prediction was obtained by averaging multi-view results produced by all the networks on different augmented versions of each image.

\paragraph{\textbf{Glaucoma classification}} Two sets of classification models were combined. One was trained using the whole fundus images, while the other was trained using only local regions around the ONH. The OD/OC area was detected using the segmentation model described above. Subsequently, the REFUGE training set was used to fine-tune pre-trained ResNet-50, -101, -152~\citep{he2016deep} and -38~\citep{wu2019wider} models. The final classification result was assigned by ensembling the outputs of these architectures by averaging. \\

\subsection*{\textbf{BUCT}}

\paragraph{\textbf{OD/OC segmentation}} The OD/OC were segmented separately by two different U-Net~\citep{ronneberger2015u} models. First, the images on the REFUGE training set were resized to fit the resolution of those on the validation set and converted to gray scale. Then, for OD segmentation, a square of $817 \times 817$ pixels was cropped from the input images, leaving the ONH on the left-hand side, and then resized to $256 \times 256$ pixels. A U-Net with less convolutional filters than the original approach~\citep{ronneberger2015u} was applied to retrieve the OD. To remove false positives, the largest connected component was taken, and an ellipse was fitted to the OD segmentation. For OC segmentation, the smallest rectangle containing the OD was clipped out, and each side of the rectangle was extended with 100 pixels to fit a resolution of $128 \times 128$ pixels. The same U-Net architecture was retrained then on these images and applied to retrieve the OC. The largest connected component was taken as the final result, too. In both cases, the U-Nets were trained using the REFUGE training set with data augmentation, including rotations and flippings.

\paragraph{\textbf{Glaucoma classification}} The same cropping strategy applied for OD/OC segmentation was used for this task. The resulting CFPs were then transformed into grayscale images. Standard data augmentation techniques such as rotations and shiftings were applied to increase the size of the training set. Then, an X-ception~\citep{chollet2017xception} network was trained from scratch for glaucoma classification using grayscale versions of the color images on the REFUGE training set and the ground truth annotations. \\

\subsection*{\textbf{CUHKMED}}

\paragraph{\textbf{OD/OC segmentation}} A patch-based Output Space Adversarial Learning framework (pOSAL)~\citep{wang2019patch} was introduced for this task. This method enables output space domain adaptation to reduce the segmentation performance degradation on target datasets with domain shift in an unsupervised way. A region of interest (ROI) containing the OD from each original image was first extracted using a U-Net~\citep{ronneberger2015u} model. The DeepLabv3+~~\citep{Chen2018} architecture was afterwards applied for segmentation, using the backbone of MobileNetV2~\citep{sandler2018mobilenetv2}. Considering the shape of the OD and OC, a morphology-aware segmentation loss was designed to force the network to generate smooth predictions. To overcome the domain shift between training and testing datasets, adversarial learning was exploited, encouraging the segmentation predictions in the target domain to be similar to the source ones. During this process, the labelled training images are considered as the source domain, while the unlabelled validation images are from the target domain. Specifically, a patch-based discriminator was introduced to distinguish whether the prediction came from the source or the target domain and the adversarial learning prompts the segmentation network to generate validation predictions similar to predictions of training images~\citep{wang2019patch}. The final image prediction was acquired by ensembling five models, to further improve the segmentation performance. Further details are provided in~\citep{wang2019patch}.

\paragraph{\textbf{Glaucoma classification}} This task was tackled without using a dedicated method. Instead, the authors proposed to use the OD/OC segmentation masks--automatically obtained with the method described above--to compute the vertical CDR (vCDR). To this end, two ellipses were fitted to the the OD and OC masks, respectively. The vCDR values were normalized into 0-1 as a final classification probability following: $p_\text{new} = \frac{p - p_\text{min}}{p_\text{max} - p_\text{min}}$, where $p$ is the calculated vCDR values, $p_\text{min}$ and $p_\text{max}$ are the minimum and maximum vCDR values among all the testing images. \\

\subsection*{\textbf{Cvblab}}

\paragraph{\textbf{OD/OC segmentation}} A two-stage process was followed for this task, based on a modified U-Net architecture~\citep{sevastopolsky2017optic}. The OD was segmented first and the resulting mask was used to crop the image and segmenting the OC. As a pre-processing technique, the Contrast Limited Adaptive Histogram Equalization (CLAHE) method, was applied. The images were also resized to $256 \times 256$ pixels before feeding the network. The models were trained using DRIONS-DB, DRISHTI-GS, RIM-ONE v3 and the REFUGE training set.

\paragraph{\textbf{Glaucoma classification}} An ensemble of VGG19~\citep{simonyan2014very}, GoogLeNet (InceptionV3)~\citep{szegedy2016rethinking}, ResNet-50~\citep{he2016deep} and the Xception~\citep{chollet2017xception} architectures was applied for this task. Each network was independently fine-tuned from the weights pre-trained from ImageNet~\citep{russakovsky2015imagenet} to identify glaucomatous images, using DRISHTI-GS1, HRF, ORIGA, RIM-ONE and the training set of the REFUGE databases. Data augmentation was applied in the form of vertical and horizontal flippings, rotations up to $50^\circ$, height/width shifts of 0.15 and zooms in a range between 0.7 and 1.3. Prior to fine tunning, the training data was balanced using SMOTE~\citep{chawla2002smote} on the REFUGE training set, with the aim of reducing the bias on the prediction model towards the more common class (Normal). All the images were resized to $256 \times 256$ pixels before feeding the network. The results were merged together by ensembling the models' outputs taking the average glaucoma likelihood. \\

\subsection*{\textbf{Mammoth}}

\paragraph{\textbf{OD/OC segmentation}} A Mask-RCNN~\citep{he2017mask} and a Dense U-Net~\citep{ronneberger2015u} were ensembled for this task. For Mask-RCNN, the OD was first segmented. Then, each input image was cropped around its center to retrieve a patch with a size of $512 \times 512$ pixels, and the segmentation of the OC was performed on it. For the Dense U-Net, which is a modified U-Net architecture with dense convolutional blocks and dilated convolutions, the OD was first segmented. Then the probability mask was used as additional channel of the input (as attention) to segment OC. Both networks were trained using a linear combination of cross-entropy and Dice losses. The probability outputs of both networks were averaged to generate the final segmentation results. A subsample from the original REFUGE training set was used to learn the models. In particular, it was divided into two new sets, one used for training (32 glaucoma images and 288 non-glaucoma images) and a second for validation (8 glaucoma images and 72 non-glaucoma images). The Mask-RCNN internally used a ResNet-50~\citep{he2016deep} model pre-trained in the COCO~\citep{lin2014microsoft} data set and fine-tuned using the above mentioned training set.

\paragraph{\textbf{Glaucoma classification}} The OD/OC segmentation method was used to crop each input image and generate a patch centered in the ONH, covering 1.5 times the radius of the OD. The resulting image was then resized to $448 \times 448$, and CLAHE contrast equalization and mean color normalization were subsequently applied to uniform image characteristics across data sets. A combination of a ResNet-18~\citep{he2016deep} (supervised) and a CatGAN~\citep{wang2017catgan} (semi-supervised) classification networks was applied for diagnosis. The CatGAN was used to aid the learning process of the ResNet-18 model in a semi-supervised setting, using fake images generated by the CatGAN to increase the size of the training set. The same training/validation partition used for OD/OC segmentation was applied for this task. A series of ResNet-18 models was trained using 4-fold cross-validation on these training set and a weighted and an unweighted cross-entropy loss, resulting in $4 \times 2 = 8$ models in total. At inference time, the predictions of all the models were averaged into a final glaucoma likelihood.  \\

\subsection*{\textbf{Masker}}

\paragraph{\textbf{OD/OC segmentation}} The first step consisted of localizing the ONH region. A Mask-RCNN~\citep{he2017mask} architecture was used to this end. Afterwards, the image was cropped around the ONH to build a new training set. This set was divided into 14 partitions based on a bagging principle. Different image preprocessing techniques were applied to each subset, namely image dehazing~\citep{berman2016non} and edge-preserving multiscale image decomposition based on weighted least squares optimization~\citep{farbman2008edge}. Different networks including Mask-RCNN~\citep{he2017mask}, U-Net~\citep{ronneberger2015u} and M-Net~\citep{fu2018joint} were trained on each subset, and the final result was obtained by a voting procedure in which regions predicted by 80$\%$ of all the networks were taken as the final segmentation.

\paragraph{\textbf{Glaucoma classification}} The vCDR value was first calculated using the segmentation results obtained with the previously described method. Subsequently, several classification networks based on ResNet~\citep{he2016deep} were trained from scratch to predict the risk of glaucoma. The REFUGE training set and ORIGA were used to learn the models. The final result was obtained based on a linear combination of the vCDR values and the prediction of the classification networks. We use ResNet-50, ResNet-101 and ResNet-152 as the basic classification models. The final glaucoma risk is:

\begin{equation}
\text{Glaucoma Risk} = 0.8 \times \text{CDR} + 0.2 \times \text{CNets}.
\end{equation}

\noindent Here, CDR is the vertical cup to disc ratio and CNets is the final voting of the ensemble classification networks. If 80$\%$ of all the networks predict a image with high risk of glaucoma, CNets = 1, otherwise, CNets = 0. In our implementation, we use 14 different networks.  \\

\subsection*{\textbf{NightOwl}}

\paragraph{\textbf{OD/OC segmentation}} A coarse to fine approach was proposed for this task, based on two dense U-shaped networks with dense blocks~\citep{huang2017densely}, namely CoarseNet (C-Net) and FineNet (F-Net), respectively. The C-Net model was used to coarsely localize the ONH region. Then, the F-Net was applied to retrieve the final segmentation of the OD and the OC. A modified version of pooling based on the mean of average and max-pooling was applied for better feature accumulation. The images were preprocessed using histogram matching to normalize the intensities in the sample space--and exponential transformations to enhance the boundaries of the optic cup---. Standard data augmentation techniques were applied to the REFUGE training set to balance the number of images from each class (glaucomatous / non-glaucomatous). The original inputs, resized to $112 \times 112$ pixels, were fed to the C-Net for localizing the ONH region. This area was then extracted from the original input image, resized to $112 \times 112$ pixels too, and fed to two different F-Nets for OD/OC segmentation. Outliers were removed using morphological operations (opening and closing) and Gaussian smoothing.

\paragraph{\textbf{Glaucoma classification}} The encoders of each F-Net were used for extracting two vectors of 2048 features each, one for the OD and one for the OC. Dimensionality reduction via convolutions was applied to retrieve two new vectors with 64 features each. The concatenation of these two vectors was used to feed a neural network with 4 fully connected layers, trained to predict the glaucoma likelihood. The weights of the F-Net encoders were not adjusted for glaucoma classification, only the weights used for dimensionality reduction and those of the fully connected layers. 10-fold cross-validation was applied to retrieve 10 different models, and 7 of them were retrieved based on their confusion matrices. The final glaucoma likelihood was obtained by taking the maximum likelihood from all the models.  \\

\subsection*{\textbf{NKSG}}

\paragraph{\textbf{OD/OC segmentation}} The DeepLabv3+~\citep{Chen2018} architecture was used for this task, based on the assumption that atrous spatial pyramid pooling (ASPP) is effective to segment objects at multiple scales. The network was trained using cross-entropy as the loss function. The images were pre-processed using pixel quantization to reduce the sensitivity of the model to changes in color and to improve its robustness. Moreover, the segmentation approach was applied on cropped versions of the input images. These were obtained by extracting a bounding box surrounding the ONH area.

\paragraph{\textbf{Glaucoma classification}} This task was performed using a SENet~\citep{Hu2018} architecture. This network has large capacity, as it has 154 layers in total. Instead of using fully connected layers, it uses $1 \times 1$ convolutions. The images were preprocessed by applying the same strategy used for segmentation. The glaucomatous/non-glaucomatous classes were balanced using re-sampling. By means of data augmentation using rotations and stretching, the REFUGE training set was increased to a total of 2000 images. \\

\subsection*{\textbf{SDSAIRC}}

\paragraph{\textbf{OD/OC segmentation}} A method inspired by the M-Net~\citep{fu2018joint} was applied for this task. An area of $480\times480$ pixels size was defined and prepared as the segmentation ROI for each image, centered on the OD and transformed to polar coordinates afterwards. The histogram of the test images were matched to the average histogram of the REFUGE training set to compensate image variance per camera vendor. The segmentation task was divided into OD segmentation from the segmentation ROI and OC segmentation from the bounding box of the OD. This box was tightly cropped to contain the entire OD. This two stage separation helped to tackle the difficulty in finding the ideal weights for the M-Net~\citep{fu2018joint}. The segmentation accuracy was further improved by post-processing the resulting masks using ellipse fitting.

\paragraph{\textbf{Glaucoma classification}} A ResNet-50~\citep{he2016deep} network with pre-trained weights from ImageNet~\citep{russakovsky2015imagenet} was fine-tuned on the REFUGE training for glaucoma classification. Histogram matching was applied to uniform the appeareance of images with respect to the training set. The CFPs were also cropped in such a way that the OD was positioned in the upper-left corner. This setting allows to capture RNFL defects in more detail than cropping a square centered in the ONH. The final glaucoma likelihood was obtained by averaging the classification score predicted by the network with the resulting score of a logistic regression which takes advantage of vCDR value, estimated from the OD/OC segmentation, as an input. To this end, the logistic regression classifier was trained separately using the transformed vCDR value. \\

\subsection*{\textbf{SMILEDeepDR}}

\paragraph{\textbf{OD/OC segmentation}} A modified U-Net~\citep{ronneberger2015u} architecture, namely X-Unet, was applied for this task. It used 3 inputs so that it was able to receive more original raw pixel information during training. This strategy was used to reduce the risk of overfitting while enhancing the network's learning capability. Moreover, squeeze-and-excitation blocks were embedded into this U-Net variant to weight the features from different convolutional layers' channels. Such a mechanism was able to selectively amplify the valuable channel-wise features and suppress the useless feature from global information. In addition, deconvolution were used in the network decoder to refine the decoding capability by refusing the features between different level encoded features and the corresponding level decoded features. The segmentation task was also posed as a linear regression task instead of a typical pixel classification problem, using $L_1$ loss for training. A \textit{split-copy-merge} strategy was followed: a X-Unet network was trained first to predict the ground labels. Secondly, two X-Unets were separately fine-tuned using the learned weights, only to predict the OD and the OC, respectively. Then, the predictions of both networks were merged to get the final result.

\paragraph{\textbf{Glaucoma classification}} The Deeplabv3+~\citep{chen2017rethinking} was modified and used as a classifier. Its last layer was replaced by a global average pooling layer followed by a fully connected layer. The model was trained on the REFUGE training set using the cross-entropy loss. Instead of using the full images, a pre-processing stage based on cropping the regions around the ONH was followed.  \\

\subsection*{\textbf{VRT}}

\paragraph{\textbf{OD/OC segmentation}} A U-Net~\citep{ronneberger2015u} based architecture was used, complemented by an auxiliary CNN~\citep{son2017retinal} that took a vessel segmentation mask and generated a coarse mask with the estimated OD/OC location. The output of the second network was concatenated to the bottleneck layer of the U-Net to generate the final segmentation mask. A combined loss $L_\text{total}=L_\text{main}+\lambda*L_\text{vessel}$ was applied, where $L_\text{main}$ and $L_\text{vessel}$ are pixel-wise binary cross entropy for the U-Net and the auxiliary CNN. The values for $\lambda$, the depth of U-Net and the number of filters at the last layer of the auxiliary CNN were experimentally selected using a hill-climbing approach. The OD and the OC were segmented separately using two different U-Net architectures. Holes in the final segmentations were filled, and the OD/OC areas were converted to convex-hulls to ensure a single binary mask per regions. 

\paragraph{\textbf{Glaucoma classification}} The method was based on three architectures as described in~\citep{son2018classification},\footnote{{\url{https://bitbucket.org/woalsdnd/refuge/src}}} each of them targetting glaucoma classification or the detection of glaucomatous disc changes and RNFL defects. The three models were trained using images from three public data sets, namely Kaggle~\citep{kaggledr}, MESSIDOR~\citep{decenciere2014feedback} and IDRiD~\citep{porwal2018indian}. Since these databases do not have labels for any of these tasks, a semi-supervised learning approach was followed. Models pre-trained on a private data set were used to assign labels to the images on each of the public sets. Given that the data sets used are still public and the assigned labels are not gold standard annotations but automated and therefore prone to errors, the organizers decided that this proposal is still in accordance with the participation rules. The same architectures used for assigning the automated labels were then trained from scratch on the combined data set to produce final predictions. The final glaucoma likelihood was assigned by doing: $\max\{ \text{glaucomatous disc change, RNFL defect} + \text{glaucoma suspect} / 2 \}$.  \\


\subsection*{\textbf{WinterFell}}

\paragraph{\textbf{OD/OC segmentation}} The ONH was initially detected using a Faster R-CNN~\citep{girshick2015fast}. This area was cropped in all the images, and two image processing techniques were applied on the outputs. The first approach consisted of selecting a standard image and then normalize the remaining ones using it as a reference. The second image version was the inverted green channel of the original RGB cropped image. Finally, a ResU-Net~\citep{shankaranarayana2017joint} model was applied on the resulting images for OD/OC segmentation. 

\paragraph{\textbf{Glaucoma classification}} An ensemble of ResNets~\citep{he2016deep} (101 and 152) and DensNets~\citep{huang2017densely} (169 and 201) was used for classification. The networks were pre-trained on ImageNet and separately fine-tuned using ORIGA, based on the log-likelihood loss. Each model was trained on cropped versions of the inputs images, centered in the ONH and on three different color spaces (RGB, HSV and the inverted green channel). Hence, $4 \times 3 = 12$ different models were produced. The final result was obtained by taking the mode of the binary decisions of each network. If the predicted label was glaucoma, the maximum confidence score was used as a final likelihood. On the contrary, if the image was labeled as non-glaucomatous, then the minimum score was applied.  \\

\end{document}

%% file: authors.tex




\author[1]{Jos\'e~Ignacio~Orlando}
\author[2]{Huazhu~Fu}
\author[3,4]{Jo\~ao~Barbossa~Breda}
\author[4]{Karel~van~Keer}
\author[5]{Deepti~R.~Bathula}
\author[6]{Andr\'es~Diaz-Pinto}
\author[7]{Ruogu~Fang}
\author[8]{Pheng-Ann~Heng}
\author[9]{Jeyoung~Kim}
\author[10]{JoonHo~Lee}
\author[10]{Joonseok~Lee}
\author[11]{Xiaoxiao~Li}
\author[7]{Peng~Liu}
\author[12]{Shuai~Lu}
\author[13]{Balamurali~Murugesan}
\author[6]{Valery~Naranjo}
\author[5]{Sai~Samarth~R.~Phaye}
\author[14]{Sharath~M.~Shankaranarayana}
\author[5]{Apoorva~Sikka}
\author[15]{Jaemin~Son}
\author[16]{Anton~van~den~Hengel}
\author[8]{Shujun~Wang}
\author[17]{Junyan~Wu}
\author[16]{Zifeng~Wu}
\author[18]{Guanghui~Xu}
\author[12]{Yongli~Xu}
\author[18]{Pengshuai~Yin}
\author[19]{Fei~Li}
\author[19]{Xiulan~Zhang}\corref{cor1}
\author[20]{Yanwu Xu}\corref{cor1}
\cortext[cor1]{Corresponding authors: Yanwu Xu (\url{ywxu@ieee.org}) and Xiulan Zhang (\url{zhangxl2@mail.sysu.edu.cn.}).}
\author[1]{Hrvoje~Bogunovi\'c}

\address[1]{Christian Doppler Laboratory for Ophthalmic Image Analysis (OPTIMA), Vienna Reading Center (VRC), Department of Ophthalmology and Optometry, Medical University of Vienna, Spitalgasse 23, 1090 Vienna, Austria.}
\address[2]{Inception Institute of Artificial Intelligence, Abu Dhabi, United Arab Emirates.}
\address[3]{Surgery and Physiology Department, Ophthalmology Unit, Faculty of Medicine, University of Porto, Porto, Portugal.}
\address[4]{Research Group Ophthalmology, KU Leuven, Leuven, Belgium}
\address[5]{Department of Computer Science \& Engineering at Indian Institute of Technology (IIT) Ropar, Rupnagar, 140001 Punjab, India.}
\address[6]{Instituto de Investigaci\'on e Innovaci\'on en Bioingenier\'ia, I3B, Universitat Polit\`ecnica de Val\`encia, 46022 Valencia, Spain.}
\address[7]{J. Crayton Pruitt Family Dept. of Biomedical Engineering, University of Florida, 32611~USA.}
\address[8]{Department of Computer Science and Engineering, The Chinese University of Hong Kong, 999077 Hong Kong.}
\address[9]{Gachon University, 461-701 Gyeonggi-do, Korea.}
\address[10]{Samsung SDS AI Research Center, 06765 Seoul, Korea.}
\address[11]{Yale University, 06510 New Haven, CT USA.}
\address[12]{Faculty of Science, Beijing University of Chemical Technology, 100029 Beijing, China.}
\address[13]{Healthcare Technology Innovation Centre, IIT-Madras,~India.}
\address[14]{Department of Electrical Engineering, IIT-Madras, India.}
\address[15]{VUNO Inc., Seoul, 137-810 Korea.}
\address[16]{Australian Institute for Machine Learning, Australia.}
\address[17]{Cleerly Inc. 10022 New York City, NY USA.}
\address[18]{South China University of Technology, 510006 Guangzhou,~China.}
\address[19]{Zhongshan Ophthalmic Center, Sun Yat-sen University, China.}
\address[20]{Artificial Intelligence Innovation Business, Baidu Inc., China and 
Cixi Institute of BioMedical Engineering, Chinese Academy of Sciences, China.}

%% file: abstract.tex
\begin{abstract}
Glaucoma is one of the leading causes of irreversible but preventable blindness in working age populations. Color fundus photography (CFP) is the most cost-effective imaging modality to screen for retinal disorders. 
However, its application to glaucoma has been limited to the computation of a few related biomarkers such as the vertical cup-to-disc ratio. 
Deep learning approaches, although widely applied for medical image analysis, have not been extensively used for glaucoma assessment due to the limited size of the available data sets. Furthermore, the lack of a standardize benchmark strategy makes difficult to compare existing methods in a uniform way.
In order to overcome these issues
we set up the Retinal Fundus Glaucoma Challenge, REFUGE (\url{https://refuge.grand-challenge.org}), held in conjunction with MICCAI 2018. The challenge consisted of two primary tasks, namely optic disc/cup segmentation and glaucoma classification. As part of REFUGE, we have publicly released a data set of 1200 fundus images with ground truth segmentations and clinical glaucoma labels, currently the largest existing one. We have also built an evaluation framework to ease and ensure fairness in the comparison of different models, encouraging the development of novel techniques in the field. 12 teams qualified and participated in the online challenge. This paper summarizes their methods and analyzes their corresponding results. In particular, we observed that two of the top-ranked teams outperformed two human experts in the glaucoma classification task.  Furthermore, the segmentation results were in general consistent with the ground truth annotations, with complementary outcomes that can be further exploited by ensembling the results.
\end{abstract}

%% file: abbreviations.tex
\section*{List of abbreviations}

\begin{itemize}
    \item abs: Absolute value.
    \item Acc: Accuracy.
    \item AMD: Age-related Macular Degeneration.
    \item ASPP: Atrous Spatial Pyramid Pooling.
    \item AUC: Area Under the (ROC) Curve.
    \item CFP: Color Fundus Photograph.
    \item CLAHE: Contrast Limited Adaptive Histogram Equalization
    \item CONV: Convolutional layer.
    \item DR: Diabetic Retinopathy.
    \item DSC: Dice coefficient.
    \item FC: Fully Connected layer.
    \item FCN: Fully Convolutional Network.
    \item FDA: US Food and Drug Administration
    \item FN: False Negatives.
    \item FOV: Field-Of-View.
    \item FP: False Positives.
    \item G: Glaucoma.
    \item HSV: Hue Saturation Value.
    \item IOP: Intra Ocular Pressure.
    \item IoU: Intersection over Union / Jaccard index.
    \item NTG: Normal Tension Glaucoma.
    \item MAE: Mean Absolute Error.
    \item MICCAI: Medical Imaging and Computer Assisted Invervention conference.
    \item OC: Optic Cup.
    \item OCT: Optical Coherence Tomography.
    \item OD: Optic Disc.
    \item ONH: Optic Nerve Head.
    \item OMIA: Ophthalmic Medical Image Analysis workshop.
    \item POAG: Primary Open Angle Glaucoma.
    \item PPA: Peripapillary Atrophy.
    \item Pr: Precision / Positive predictive value.
    \item REFUGE: Retinal Fundus Glaucoma challenge.
    \item RGB: Red Green Blue.
    \item RNFL: Retinal Nerve Fiber Layer.
    \item ROC: Receiver-Operating Characteristic curve.
    \item ROI: Region Of Interest.
    \item Se: Sensitivity.
    \item SMOTE: Synthetic Minority Oversampling Technique.
    \item Sp: Specificity / True negative ratio.
    \item TN: True Negatives.
    \item TP: True Positives.
    \item vCDR: Vertical Cup-to-Disc Ratio.
\end{itemize}

%% file: introduction.tex
\section{Introduction}

Glaucoma is a chronic neuro-degenerative condition that is one of the leading causes of irreversible but preventable blindness in the world~\citep{tham2014global}. In 2013, 64.3 million people aged 40-80 years were estimated to suffer from glaucoma, while this number is expected to increase to 76 million by 2020 and 111.8 million by 2040~\citep{tham2014global}. In its many variants, glaucoma is characterized by the damage of the optic nerve head (ONH), typically caused by a high intra-ocular pressure (IOP). IOP is increased as a consequence of abnormal accumulation of aqueous humor in the eye, induced by pathological defects in the eye's drainage system. When the anterior segment is saturated with this fluid, the IOP progressively elevates, compressing the vitreous to the retina. If this remains uncontrolled, it can produce damage in the nerve fiber layer, the vasculature and the ONH, leading to a progressive and irreversible vision loss that can ultimately result in blindness. As this process occurs asymptomatically, glaucoma is frequently referred as the \textit{"silent thief of sight"}~\citep{schacknow2010glaucoma}: patients are not aware of the progressing disease until the vision is irreversibly lost.

Life-long pharmacological treatments based on the regular administration of eye drops are usually prescribed to control the IOP and to temper further damage in the retina. Alternatively, laser procedures and other surgeries can be performed to increase the drainage. In any case, early detection is essential to prevent vision loss~\citep{schacknow2010glaucoma}. Unfortunately, at least half of patients with glaucoma currently remain undiagnosed~\citep{ProkofyevaZrenner2012}. Being glaucoma a chronic condition, one of the major challenges is to be able to detect this large number of undiagnosed patients~\citep{ProkofyevaZrenner2012}. Generalized screening programs have not been employed because of the large amount of false positives these can generate. These misdiagnoses cannot be absorbed by current healthcare infrastructures and would have an unnecessary negative impact on the patient's quality of life, until it would be recognized that no glaucomatous neuropathy existed~\citep{schacknow2010glaucoma}. 

\begin{figure}[!t]
\centering
\includegraphics[width=0.7\textwidth]{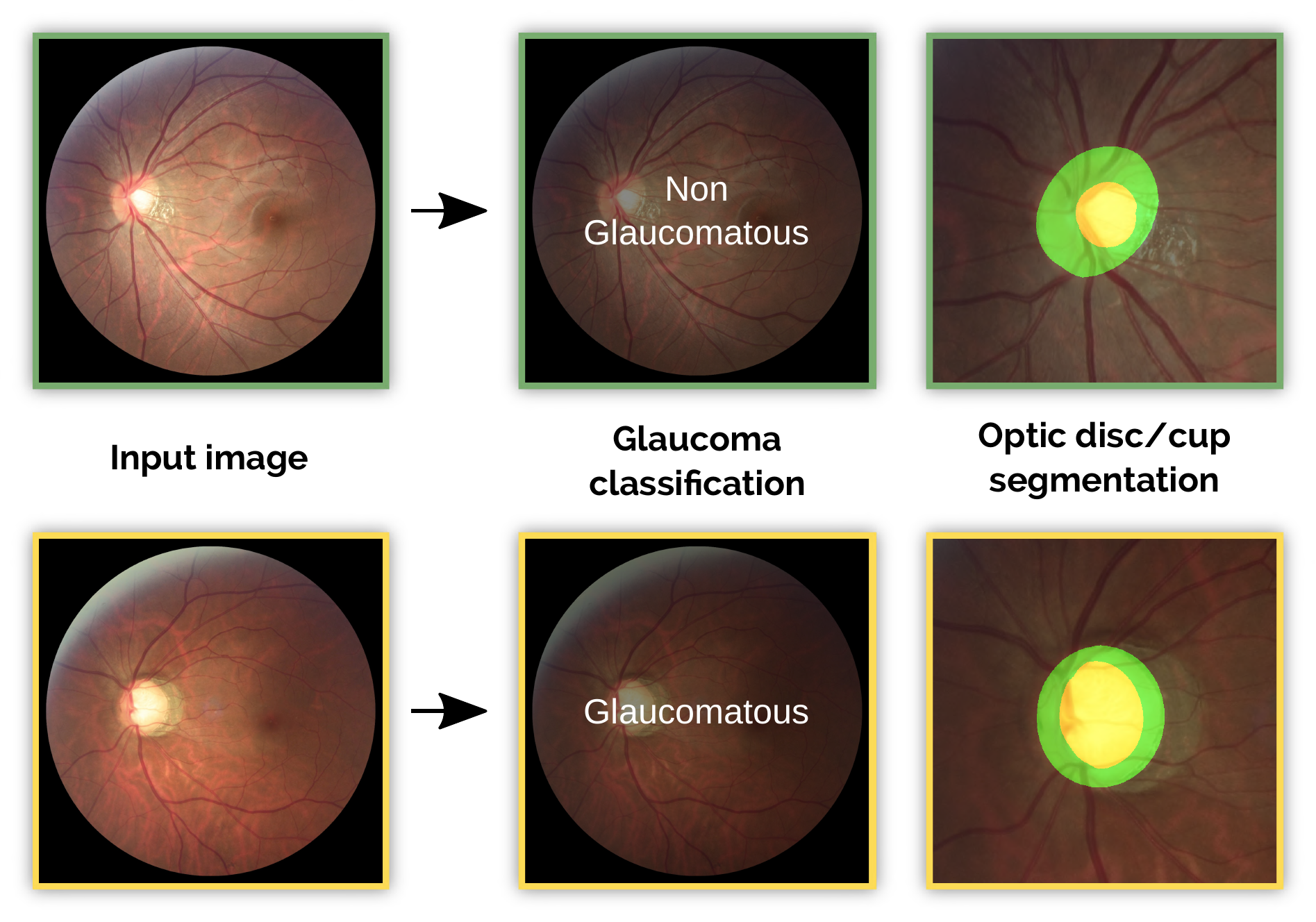}
\caption{REFUGE challenge tasks: glaucoma classification and optic disc/cup segmentation from color fundus photographs.}
\label{fig:challenge-description}
\end{figure}

\begin{figure}[!t]
\centering
\includegraphics[width=1.0\textwidth]{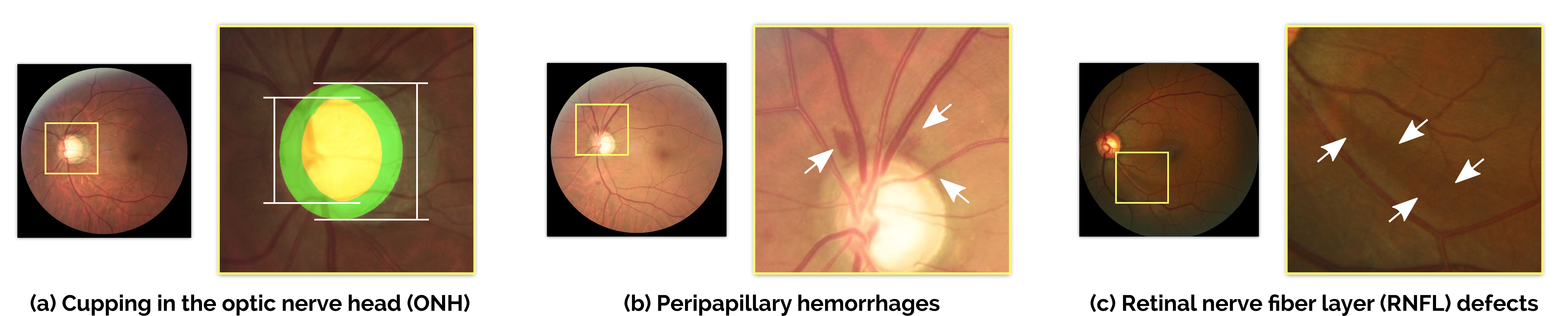}
\caption{Pathological changes typical from glaucoma, as observed through fundus photography. (a) Neuroretinal rim thinning due to cupping in the optic nerve head (ONH). White lines indicate the vertical diameter of the optic disc (green) and the optic cup (yellow). (b) Peripapillary hemorrhages, observed as flame-shaped bleedings in the vicinity of the ONH. (c) Retinal nerve fiber layer defects are observed as subtle striations spanning from the optic disc border.}
\label{fig:glaucoma-detection-challenging}
\end{figure}

Color fundus photography (CFP, Figure~\ref{fig:challenge-description}) is currently the most economical, non-invasive imaging modality for inspecting the retina~\citep{abramoff2010retinal,SE2018ai}. Its widespread availability makes it ideal for assessing several ophthalmic diseases such as age-related macular degeneration (AMD)~\citep{burlina2017automated}, diabetic retinopathy (DR)~\citep{gulshan2016development} and glaucoma~\citep{li2018efficacy}. Screening campaigns can be aided by the incorporation of computer-assisted tools for image-based diagnosis. As these initiatives require to manually grade a large number of cases in a short period of time, automated tools can help clinians by providing them with quantitative and/or qualitative feedback (e.g. disease likelihood, segmentations of relevant lesions and pathological structures, etc). These approaches have already been successfully applied for detecting DR, in a FDA-approved autonomous diagnostic system, a first of its kind~\citep{abramoff2018pivotal}.
However, the broad application of similar methods for glaucoma detection is still pending. This is partially due to the fact that the earlier signs of glaucoma are not so easily recognizable in CFP~\citep{lavinsky2017future} (Figure~\ref{fig:glaucoma-detection-challenging}). In current best clinical practice, CFPs are complementary to other studies such as IOP measurements, automated perimetry and optical coherence tomography (OCT). This approach is not cost-effective to be applied for large scale population screening for glaucoma~\citep{schacknow2010glaucoma}. Therefore, developing automated tools to better exploit the information in CFP is paramount to reduce this burden and ensure an effective detection of glaucoma suspects.

A significant research effort has been made to introduce automated tools for segmenting the optic disc (OD) and the optic cup (OC) in CFP automatically, or to identify glaucomatous cases based on alternative features~\citep{almazroa2015optic,haleem2013automatic,thakur2018survey}. Nevertheless, these approaches currently cannot be properly compared due to the lack of a unified evaluation framework to validate them. Moreover, the absence of large scale public available data sets of labeled glaucomatous images has hampered the rapid deployment of deep learning techniques for glaucoma detection~\citep{hagiwara2018computer}. It has been recently shown that image analysis competitions in general can aid to identify challenging scenarios that need further development~\citep{prevedello2019challenges}. Recent grand challenges such as ROC~\citep{niemeijer2010retinopathy}, Kaggle~\citep{kaggledr} and IDRiD~\citep{porwal2018indian}, on the other hand, have shown to be useful to address both inconveniences in DR~\citep{SE2018ai}, favoring the deployment of these tools into the daily clinical practice~\citep{abramoff2018pivotal}. Unfortunately, similar initiatives have not been introduced for glaucoma detection and/or assessment yet.

In an effort to overcome these limitations, we introduced the Retinal Fundus Glaucoma Challenge (REFUGE), a competition that was held as part of the Ophthalmic Medical Image Analysis (OMIA) workshop at MICCAI 2018. The key contributions of the challenge were: (i) the release of a large database (approximately two times bigger than the largest available so far) of 1200 CFP with reliable reference standard annotations for glaucoma identification, optic disc/cup (OD/OC) segmentation and fovea localization; and (ii) the constitution of a unified evaluation framework that enables a standardized fair protocol to compare different algorithms. To the best of our knowledge, REFUGE is the first initiative to provide these key tools at such a large scale. REFUGE participants were invited to use the data set to train and evaluate their algorithms for glaucoma classification and OD/OC segmentation. Their results were quantitatively evaluated using our uniform protocol, to ensure a fair comparison.

In this paper, we analyze the outcomes and the methodological contributions of REFUGE. We present and describe the challenge, reporting the performance of the best algorithms evaluated in the competition and identifying successful common practices for solving the proposed tasks. The results are contrasted with the outcomes of two glaucoma experts to study their performance with respect to independent human observers. Finally, we take advantage of all these empirical evidence to discuss the clinical implications of the results and to propose further improvements to this evaluation framework. In line with the recommendations of~\cite{trucco2013validating}, REFUGE data and evaluation remain open to encourage further developments and ensure a proper and fair comparison of those new proposals.

%% file: previous_works.tex
\section{Automated glaucoma assessment: state-of-the-art and current evaluation protocols}
\label{sec:previous-works}

Early attempts for glaucoma classification and OD/OC segmentation were mostly based on hand-crafted methods using a combination of feature extraction techniques and supervised or unsupervised machine learning classifiers~\citep{almazroa2015optic,haleem2013automatic,thakur2018survey}. However, their accuracy was limited due to the application of manually designed features, which are unable to comprehensively characterize the large variability of disease appearance. Deep learning techniques, on the contrary, automatically learn these characteristics by exploiting the implicit information of large training sets of annotated images~\citep{litjens2017survey}. In this section we briefly analyze the state-of-the-art techniques for glaucoma classification and OD/OC segmentation and their main evaluation issues.
The interested reader could refer to the surveys by \cite{almazroa2015optic}, \cite{haleem2013automatic} and \cite{thakur2018survey} for a comprehensive analysis of the previous non-deep learning based approaches.

\subsection{Glaucoma classification}

Glaucoma classification consists in categorizing an input CFP into glaucomatous or non-glaucomatous, based on its visual characteristics.
A summary table of the most recent deep learning methods introduced for this task is available in the Supplementary Materials.
In general, most of the existing approaches are based on adaptations of standard deep supervised learning techniques, customized to deal with small training sets (Section~\ref{subsec:evaluation-protocols}). 
\cite{chen2015glaucoma}, \cite{chen2015automatic} and \cite{raghavendra2018deep} proposed to use shallow architectures with a limited number of layers. This is useful to prevent overfitting but limits the ability of the networks to learn rare, specific features.
Alternatively, the studies by \cite{christopher2018performance}, \cite{li2018combining} and \cite{orlando2017convolutional} used transfer learning methods, based on deeper architectures but pre-trained on non-medical data. \cite{christopher2018performance} fine-tuned a network initialized with weights learned from ImageNet~\citep{russakovsky2015imagenet} to detect glaucomatous optic neuropathy. Similarly, transfer learning was shown by~\cite{gomez2019automatic} to outperform networks trained from scratch for glaucoma detection. Both studies applied a massive image data set with more than 14.000 images to fine tune these networks. Other works such as those by \cite{orlando2017convolutional} and \cite{li2018combining} used deep learning features extracted from the last fully connected layers of pre-trained networks. The classification task was then performed using linear classifiers trained with these features~\citep{li2018combining,orlando2017convolutional}. This allows to use smaller data sets, although at the cost of lower performance.


Another widely used approach is to restrict the area of analysis to the ONH. This region is the one that is mostly affected by glaucoma, and focusing only there allows for a better exploitation of model parameters. This was done by most of the surveyed methods (as observed in Table~1 from the Supplementary Materials) and it resulted in a better performance than when learning from full size images. However, such a strong restriction in the networks' field of view hampers their ability to learn alternative features from other regions~\citep{chen2015glaucoma}.

\subsection{Optic disc/cup segmentation}

Segmenting the OD and the OC from CFPs is a challenging but relevant task that helps to assess glaucomatous damage to the ONH~\citep{haleem2013automatic}. Automated methods have to be robust against complex pathological changes such as peripapillary atrophies (PPA) or hemorrhages~\citep{almazroa2015optic,thakur2018survey} (Figure~\ref{fig:glaucoma-detection-challenging} (b)). On the other hand, the accurate delineation of the OC is specially difficult due to the high vessel density in the area and the lack of depth information in CFP~\citep{miri2015multimodal}. Alternative features such as vessels bendings~\citep{joshi2011optic} or intensity changes~\citep{xu2014optic} have been studied in the past to approximate the ONH depth. 
The interested reader could refer to Table 2 from the Supplementary Materials for a summary of current deep learning approaches for simultaneous OD/OC segmentation.

Most of existing methods use a surrogate segmentation/detection approach to first localize the ONH area and them crop the images around it~\citep{edupuganti2018automatic,fu2018joint,lim2015integrated,sevastopolsky2017optic,zilly2015boosting}. This prevents false positive detections in regions with e.g. severe illumination artifacts and grants a better exploitation of model parameters, as they are only dedicated to characterize the local appeareance of the OD/OC and not to differentiate these structures from other fundus regions. Alternatively, a two-stage approach was followed by \cite{sevastopolsky2018stack}, using a first neural network to retrieve a coarse segmentation and a second one to refine the results. 

Different neural network architectures have been proposed for OD/OC segmentation. \cite{lim2015integrated} applied a classification network similar to LeNet~\citep{lecun1998gradient} at a patch level to classify its central pixel as belonging to the OD, the OC or the background. 
Using patches as training samples artificially increases the available training data, although at the cost of loosing spatial information. Alternatively, Zilly \textit{et al.} proposed to overcome the data limitation issue by training a convolutional neural network using an entropy sampling approach instead of gradient descent. Most of the recent methods~\citep{al2018dense,edupuganti2018automatic,fu2018joint,sevastopolsky2017optic,sevastopolsky2018stack}, however, are based on modifications to the original U-Net architecture~\citep{ronneberger2015u}. This is due to the fact that this network can achieve good results even when trained using a relatively small amount of images. Architecture changes that heavily increase the capacity of the networks such as those introduced by \cite{edupuganti2018automatic} usually demand the application of transfer learning in the encoding path. In addition, heavy data augmentation through different combination of image transformations has also been explored \citep{fu2018joint,sun2018localizing}.

\begin{table*}
\centering
\caption{Comparison of the REFUGE challenge data set with other publicly available databases of color fundus images. Question marks indicate missing information, and N/A stands for "not applicable".}
\resizebox{\textwidth}{!}{
\begin{tabular}{C{3cm}|C{2cm} C{2cm}|C{1.5cm}| C{1.75cm} C{1.75cm} C{1.2cm}|C{2cm}|C{2cm}|C{2cm}|C{2cm}}
  \hline
  \multirow{2}{3cm}{\centering \textbf{Dataset}} & \multicolumn{3}{c|}{\textbf{Num. of images}} & \multicolumn{3}{c|}{\textbf{Ground truth labels}} & \multirow{2}{2cm}{\centering \textbf{Different cameras}} & \multirow{2}{2cm}{\centering  \textbf{Training \& test split}} & \multirow{2}{2cm}{\centering  \textbf{Diagnosis from}} & \multirow{2}{2cm}{\centering \textbf{Evaluation framework}} \\
  \cline{2-7}
  & \textbf{Glaucoma} & \textbf{Non glaucoma} & \textbf{Total} & \textbf{Glaucoma classification} & \textbf{Optic disc/cup (assessed on CFP)} & \textbf{Fovea localization} & & & & \\
  \hline
  ARIA~\citep{zheng2012automated} & 0 & 143 & 143 & No & Yes/No & Yes & No & No & ? & No \\
  \hline
  DRIONS-DB~\citep{carmona2008identification} & - & - & 110 & No & Yes/No & No & ? & No & N/A & No \\
  \hline
  DRISHTI-GS \citep{sivaswamy2014drishti,sivaswamy2015comprehensive} & 70 & 31 & 101 & Yes & Yes/Yes & No & No & Yes & Image & No \\
  \hline
  DR HAGIS~\citep{holm2017dr} & 10 & 29 & 39 & Yes & No/No & No & Yes & No & Clinical & No \\
  \hline
  IDRiD~\citep{porwal2018indian} & 0 & 516 & 516 & No & Yes/No & Yes & No & Yes & ? & Yes \\
  \hline
  HRF~\citep{odstrcilik2013retinal} & 15 & 30 & 45 & Yes & No/No & No & No & No & Clinical & No \\
  \hline
  LES-AV~\citep{orlando2018towards} & 11 & 11 & 22 & Yes & No/No & No & No & No & Clinical & No \\
  \hline
  ONHSD~\citep{lowell2004optic} & - & - & 99 & No & Yes/No & No & No & No & N/A & No \\
  \hline
  ORIGA~\citep{zhang2010origa} & 168 & 482 & 650 & Yes & Yes/Yes & No & ? & No & ? & No \\
  \hline
  RIM-ONE~\citep{fumero2011rim} v1 & 40 & 118 & 158 & Yes & Yes/No & No & No & No & Clinical & No \\
  \hline
  RIM-ONE~\citep{fumero2011rim} v2 & 200 & 255 & 455 & Yes & Yes/No & No & No & No & Clinical & No \\
  \hline
  RIM-ONE~\citep{fumero2011rim} v3 & 74 & 85 & 169 & Yes & Yes/No & No & No & No & Clinical & No \\
  \hline
  RIGA~\citep{almazroa2018retinal} & - & - & 750 & No & Yes/Yes & No & Yes & No & ? & No \\
  \hline
  \hline
  \textbf{REFUGE} & \textbf{120} & \textbf{1080} & \textbf{1200} & \textbf{Yes} & \textbf{Yes/Yes} & \textbf{Yes} & \textbf{Yes} & \textbf{Yes} & \textbf{Clinical} & \textbf{Yes} \\
  \hline
\end{tabular}}
\label{table:database-comparison} 
\end{table*}

\subsection{Evaluation protocols}
\label{subsec:evaluation-protocols}


Large discrepancies in the evaluation protocols were observed in the surveyed literature, regardless of the target task. These differences (summarized in Tables 1 and 2 of the Supplementary Materials), are mostly related with two key aspects: (i) the data sets used for training/evaluation, and (ii) the evaluation metrics.

\subsubsection{Data sets}

Table~\ref{table:database-comparison} summarizes the public available data sets of CFPs for glaucoma classification and/or OD/OC segmentation used by the literature. The REFUGE database (Section~\ref{subsec:refuge-database}) is included for comparison purposes.

In general, we observed that a lack of pre-defined partitions into training and test sets has induced a chaotic practical application of the existing data. As discussed by \cite{trucco2013validating}, this affect the feasibility of directly comparing the performance of existing methods, difficulting to conclude which model characteristics are more appropriate to solve each task. To the best of our knowledge, DRISHTI-GS\footnote{\url{http://cvit.iiit.ac.in/projects/mip/drishti-gs/mip-dataset2/Home.php}}~\citep{sivaswamy2014drishti,sivaswamy2015comprehensive} is the only existing database for glaucoma assessment that provides a clear training/test split. 

Another important aspect is related with the reliability of the assigned diagnostic labels. 
Some public data sets such as DRISHTI-GS provide glaucoma labels that were assigned based only on image characteristics. This has been also observed in private data sets such as those used by \cite{christopher2018performance} and \cite{li2018efficacy}, which were built using images from Internet that were manually graded based on their visual appeareance, without additional clinical information. Surprisingly, no information about the source of the diagnostic labels is provided in most of existing databases (see Table~\ref{table:database-comparison}). Using images with labels that were not assigned using retrospective analysis of clinical records can be problematic as it might bias automated methods to reproduce wrong labelling practices. On the contrary, clinical labels can aid algorithms to learn and discover other supplemental manifestations of the disease that are still unknown or that are too difficult to distinguish with the naked eye.

The amount of images and their diversity is also an important aspect to consider. In particular, existing databases rarely include images obtained from different acquisitions devices, ethnicities or presenting challenging glaucoma related scenarios. Therefore, the learned models might exhibit a weak generalization ability. To partially bypass this issue, some authors have proposed to train their methods using combinations of different data sets~\citep{cerentinia2018automatic,pal2018g}. 

As indicated in Table~\ref{table:database-comparison}, all existing data sets with OD/OC annotations contain manually assigned labels obtained from the CFP, without considering depth information and performed by a single reader. Consequently, these segmentations might suffer from deviations that could bias the subsequent evaluations. Incorporating depth information e.g. through stereo imaging or OCT would ensure much trustworthy annotations. On the other hand, providing segmentations obtained by the consensus of multiple readers could better approximate the true anatomy by reducing inter-observer variability.

Finally, it is important to highlight the lack of a large public data set providing both OD/OC segmentations and clinical diagnostics simultaneously. 
ONHSD\footnote{\url{http://www.aldiri.info/Image\%20Datasets/ONHSD.aspx}} \citep{lowell2004optic} and  DRIONS-DB\footnote{\url{http://www.ia.uned.es/~ejcarmona/DRIONS-DB.html}} \citep{carmona2008identification} only include segmentations of the OD, and no glaucoma labels are given. ARIA\footnote{\url{https://eyecharity.weebly.com/aria_online.html}} \citep{zheng2012automated} provides OD segmentations and incorporates vessel segmentations and annotations of the fovea center. However, the images correspond to normal subjects and patients with DR and AMD, and no segmentations of the OC are included.
DR HAGIS\footnote{\url{https://personalpages.manchester.ac.uk/staff/niall.p.mcloughlin/}} \citep{holm2017dr}, HRF\footnote{\url{https://www5.cs.fau.de/research/data/fundus-images/}} \citep{odstrcilik2013retinal} and LES-AV\footnote{\url{https://ignaciorlando.github.io/data/LES-AV.zip}} \citep{orlando2018towards}, on the other hand, include reliable diagnostic labels and vessel segmentations, but no labels for the OD/OC. Moreover, their size is relatively small (39, 45 and 22 images, respectively). RIGA\footnote{\url{https://deepblue.lib.umich.edu/data/concern/data_sets/3b591905z}} \citep{almazroa2018retinal} is a recent data set that contains 750 fundus images with OD/OC segmentations but without glaucoma labels. The three releases of RIM-ONE (v1, v2 and v3)~\citep{fumero2011rim} provide image-level glaucoma labels and OD segmentations. RIM-ONE v1 and v2 include CFPs cropped around the ONH. Furthermore, RIM-ONE v1 incorporate OD annotations by five different experts and image level labels for control subjects, ocular hypertensive patients and subjects with early, moderate and deep glaucoma. RIM-ONE v2 and v3, on the contrary, only include OD segmentations by two experts, and the diagnostic labels are classified into normal and glaucoma suspect cases. Moreover, RIM-ONE v3 do not include typical CFPs but stereo images. To the best of our knowledge, only DRISHTI-GS and ORIGA \citep{zhang2010origa} include both glaucoma classification labels and OD/OC segmentations. 
The diagnostic labels in DRISHTI-GS, however, were assigned solely based on the images~\citep{sivaswamy2015comprehensive}. ORIGA, on the other hand, is not publicly available anymore.



\subsubsection{Metrics} 

Most of the literature in glaucoma classification uses receiver-operating characteristic (ROC) curves~\citep{davis2006relationship} for evaluation, including the area under the curve (AUC) as a summary value~\citep{chen2015glaucoma,chen2015automatic,christopher2018performance,fu2018joint,gomez2019automatic,orlando2017convolutional,li2018combining,li2018efficacy,liu2018deep,pal2018g}. Sensitivity and specificity~\citep{chen2015automatic,christopher2018performance,fu2018joint,gomez2019automatic,li2018combining,liu2018deep} are also used in different studies to complement the AUC when targetting binary classification outcomes. Accuracy was reported in~\citep{cerentinia2018automatic,raghavendra2018deep} as another evaluation metric, although this metric might be biased if the proportion of non-glaucomatous images is significantly higher than the glaucomatous ones~\citep{orlando2017discriminatively}. To overcome this limitation, \cite{fu2018joint} used a balanced accuracy, consisting on the average between sensitivity and specificity.

Current literature in OD/OC segmentation make use of classical overlap metrics such as the intersection-over-union (IoU, also known as Jaccard index)~\citep{al2018dense,edupuganti2018automatic,fu2018joint,lim2015integrated,sevastopolsky2017optic,sevastopolsky2018stack,sun2018localizing,zilly2015boosting} and the Dice index~\citep{,al2018dense,edupuganti2018automatic,sevastopolsky2017optic,sevastopolsky2018stack,sun2018localizing,zilly2015boosting}. Although different by definition, these two metrics can be computed from each other, as they are defined as ratios of overlap between the predicted area and the manual reference annotation~\citep{taha2015metrics}. Pixelwise sensitivity and specificity values have been also reported in~\citep{al2018dense,fu2018joint} to illustrate the behavior in terms of false negatives and false positives, respectively. Finally, the accuracy for segmenting both the OD and the OC has been simultaneously assessed by means of the mean absolute error (MAE) of the estimated vs. manually graded CDR values~\citep{fu2018joint,lim2015integrated,sun2018localizing}. 

All these metrics are well-known and were previously used in several domains. However, it is still necessary to come up with a uniform evaluation criteria to assist method comparison and prevent the usage of potentially biased metrics.

%% file: evaluation_framework.tex
\section{The REFUGE challenge}
\label{sec:evaluation-framework}


This section briefly describes REFUGE challenge, introducing the released data set (Section~\ref{subsec:refuge-database}) and the proposed evaluation procedure (Section~\ref{subsec:challenge-setup}).

\subsection{REFUGE database}
\label{subsec:refuge-database}

\begin{figure}[t]
\centering
\includegraphics[width=0.5\textwidth]{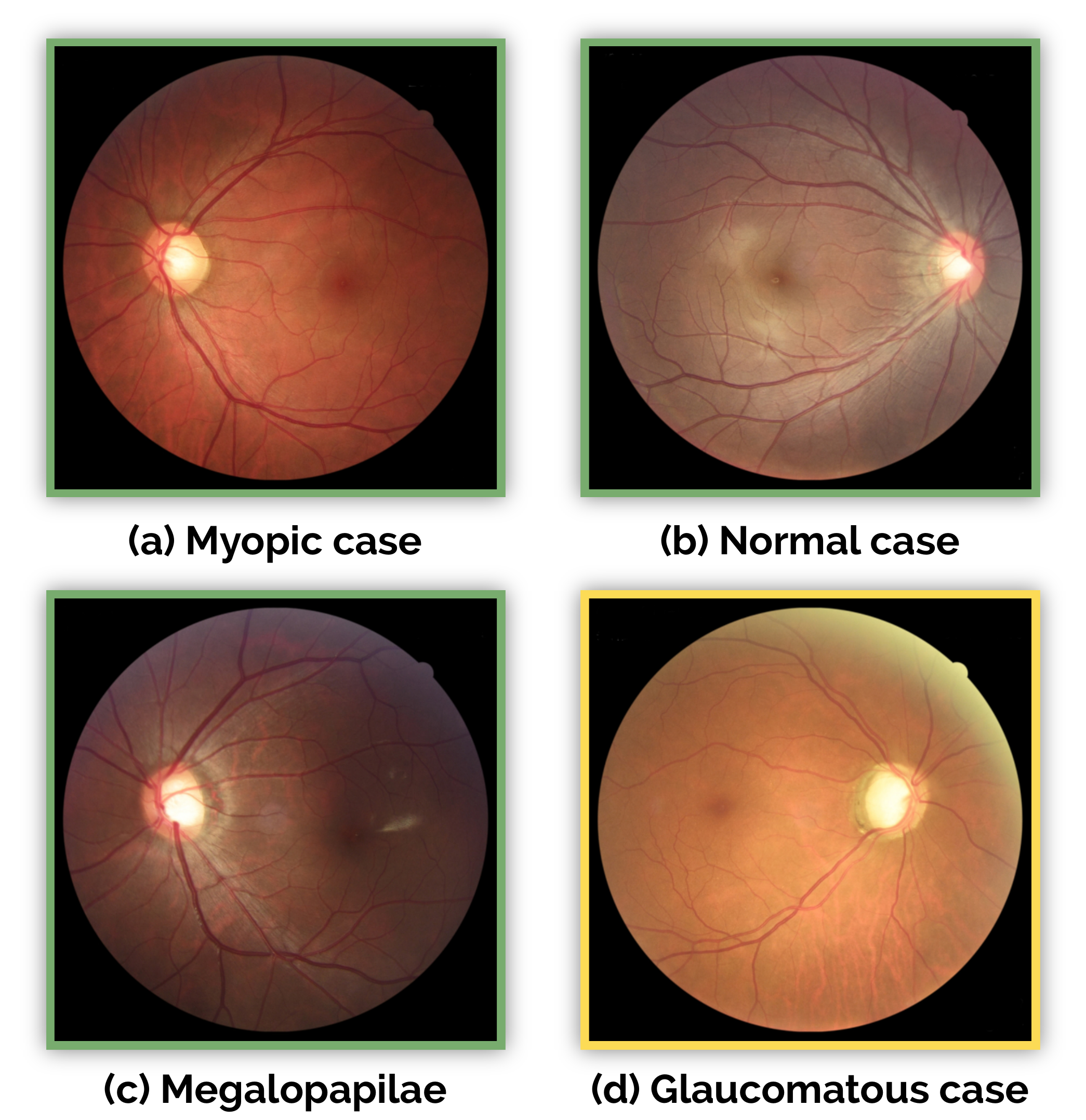}
\caption{Representative examples of color fundus photographs from the REFUGE data set. Non-glaucomatous (green) and glaucomatous (yellow) groups. (a) Myopic case with enlarged optic cup. (b) Healthy subject. (c) Patient with megalopapilae. (d, yellow) Glaucomatous case with cupping.}
\label{fig:representative-examples}
\end{figure}

The REFUGE challenge database consists of 1200 retinal CFPs stored in JPEG format, with 8 bits per color channel, acquired by ophthalmologists or technicians from patients sitting upright and using one of two devices: a Zeiss Visucam 500 fundus camera with a resolution of $2124 \times 2056$ pixels (400 images) and a Canon CR-2 device with a resolution of $1634 \times 1634$ pixels (800 images). The images are centered at the posterior pole, with both the macula and the optic disc visible, to allow the assessment of the ONH and potential retinal nerve fiber layer (RNFL) defects. These pictures correspond to Chinese patients (52\% and 55\% female in offline and online test sets, respectively) visiting eye clinics, and were retrieved retrospectively from multiple sources, including several hospitals and clinical studies. Only high-quality images were selected to ensure a proper labelling, and any personal and/or device information was removed for anonymization.

Each image in the REFUGE data set includes a reference, trustworthy glaucomatous / non-glaucomatous label. These diagnostics were assigned based on the comprehensive evaluation of the subjects' clinical records, including follow-up fundus images, IOP measurements, optical coherence tomography images and visual fields (VF). The glaucomatous cases correspond to subjects with glaucomatous damage in the ONH area and reproducible glaucomatous VF defects. This last characteristic was defined as a reproducible reduction in sensitivity compared to the normative data set, in reliable tests, at: (1) two or more contiguous locations with $p\text{-value} < 0.01$ and (2) three or more contiguous locations with $p\text{-value} < 0.05$. ONH damage was defined as a vCDR $> 0.7$, thinning of the RNFL, or both, without a retinal or neurological cause for VF loss. Notice, then, that instead of using labels assigned based on a single CFP at a specific timepoint, the labels were retrieved from examinations of follow-up medical records using a pre-determined criterion, to ensure the reliability of the classification labels.
10\% of the dataset (120 samples) corresponds to glaucomatous subjects, including Primary Open Angle Glaucoma (POAG) and Normal Tension Glaucoma (NTG). This proportion of diseased cases deviates from the global prevalence of glaucoma ($\approx4~\%$ for populations aged 40-80 years~\citep{tham2014global}). However, reducing the size of the glaucoma set would have negatively affected the ability of the classification approaches to learn features from the diseased cases. Furthermore, in an effort to model a more representative clinical scenario, the non-glaucomatous set was designed to include not only normal healthy cases but also patients with non-glaucomatous conditions such as diabetic retinopathy, myopia and megalopapilae. Myopic and megalopapilae cases were included as subjects suffering from them can easily be missclassified as glaucomatous due to their aberrant ONH appeareance (Figure~\ref{fig:representative-examples}).

Manual annotations of the OD and the OC were provided by seven independent glaucoma specialists from the Zhongshan Ophthalmic Center (Sun Yat-sen University, China), with an average experience of 8 years in the field (ranging from 5 to 10 years). All the ophthalmologists independently reviewed and delineated the OD/OC in all the images, without having access to any patient information or knowledge of disease prevalence in the data. The annotation procedure consisted in manually drawing a tilted ellipse covering the OD and the OC, separately, by means of a free annotation tool with capabilities for image review, zoom and ellipse fitting. A single segmentation per image was afterwards obtained by taking the majority voting of the anotations of the seven experts. A senior specialist with more than 10 years of experience in glaucoma performed a quality check afterwards, analyzing the resulting masks to account for potential mistakes. When errors in the annotations were observed, this additional reader analyzed each of the seven segmentations, removed those that were considered failed in his/her opinion and repeated the majority voting process with the remaining ones. Only a few cases had to be corrected using this protocol.

Manual pixel-wise annotations of the fovea were also assigned to the images to complement the data set. The fovea position was fixed by the seven independent glaucoma specialists, and a reference standard was created taking the average of these annotations.

The entire set was divided into three fixed subsets: training, offline and online test sets, each of them stratified in such a way that they contain an equal proportion of glaucomatous (10\%) and non-glaucomatous (90\%) cases. Table~\ref{table:data-characteristics} summarize the main characteristics of each subset. The training set contains all the images acquired with the Zeiss Visucam 500 camera, while the offline and online test sets include the lower resolution images captured with the Canon CR-2 device. This was made on purpose to encourage the teams to develop tools with enough generalization ability to deal with images acquired with at least using two different devices and at two different resolutions.

\begin{figure}[!t]
\centering
\includegraphics[width=0.9\columnwidth]{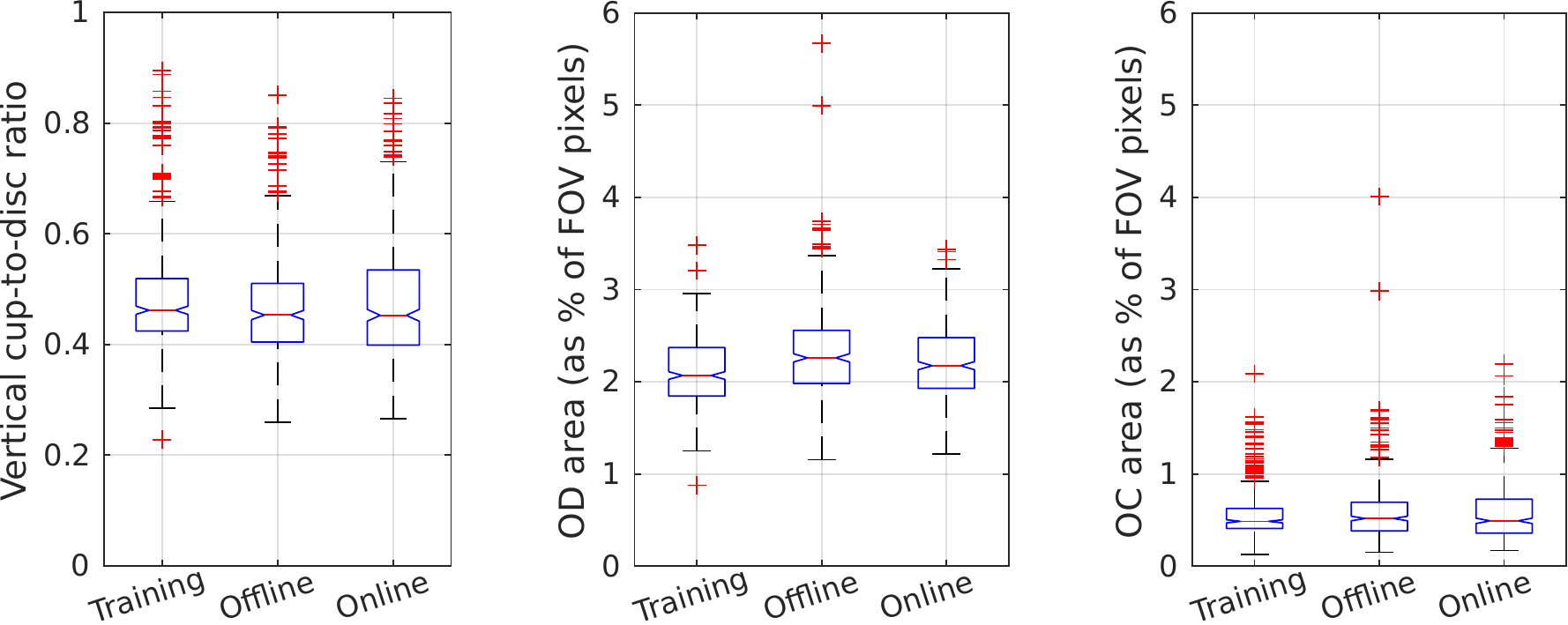}
\caption{REFUGE data set characteristics in each of the challenge partitions (training set, offline test set and online test set). From left to right: vertical cup-to-disc ratio (vCDR) values, and optic disc and cup areas, as percentages of the field-of-view area.}
\label{fig:dataset-characteristics}
\end{figure}

Figure~\ref{fig:dataset-characteristics} represents the distribution of vCDR and OD and OC areas of the images within each subset. To account for the differences in the field-of-view (FOV) of acquisitions from the Zeiss and Canon devices, the areas (in pixels) were normalized as a proportion of the FOV area (in pixels). The differences between groups were statistically assessed using Kruskal-Wallis tests with $\alpha=0.01$. Statistical significant differences were only observed for the OD area ($p = 1.4 \times 10^{-7}$, explained by the training set having larger values than the offline and online test sets ($p < 0.0091$, two-tailed Wilcoxon rank sum tests with a Bonferroni corrected significance $\alpha=0.025$ to account for the two comparisons).

\begin{table}[t]  
\centering
\caption{Summary of the main characteristics of each subset of the REFUGE data set.}
\begin{tabular}{c|C{3cm}|C{3cm}|C{3cm}}
 \hline
 {\multirow{2}{*}{\textbf{Characteristics}}} & \multicolumn{3}{|c}{\textbf{Subset}} \\
 \cline{2-4}
  & \textbf{Training} & \textbf{Offline test set} & \textbf{Online test set}\\
 \hline
 {\textbf{Acquisition device}} & Zeiss Visucam 500 & \multicolumn{2}{c}{Canon CR-2} \\
 \hline
 {\textbf{Resolution}} & $2124 \times 2056$ & \multicolumn{2}{c}{$1634 \times 1634$} \\
 \hline
 {\textbf{Num. images}} & 400 & 400 & 400 \\
 \hline
 {\textbf{Glaucoma/Non glaucoma}} & 40/360 & 40/360 & 40/360 \\
 \hline
 {\textbf{Public labels?}} & \cmark & \xmark & \xmark \\ 
 \hline
\end{tabular}
\label{table:data-characteristics} 
\end{table}

\subsection{Challenge Setup, Evaluation Metrics and Ranking Procedure}
\label{subsec:challenge-setup}

REFUGE was held in conjunction with the 5th Ophthalmic Medical Image Analysis (OMIA) workshop, during MICCAI 2018 (Granada, Spain). The challenge proposal was accepted after assessing the compliance to good practices proposed in \citep{lena2018winner,reinke2018exploit}. Thereafter, REFUGE was announced in several platforms to maximize its visibility, including the MICCAI website, its associated mailing lists and on the Grand Challenges in Biomedical Image Analysis website.~\footnote{\url{grand-challenge.org}} The challenge was officially launched in June 2018 by releasing the training set (images and labels) on a dedicated website (\url{https://refuge.grand-challenge.org/Home/}). The registered teams were allowed to use the training set to learn and adjust their proposed algorithms for glaucoma classification, OD/OC segmentation and, optionally, for fovea detection. We will not focus on this last task as it was not mandatory for participating on the challenge, and therefore no team submitted results for it on the test set. 
The registered teams were allowed to use any other public data set for developing their methods, provided that they were easily accessible by everyone. 

The offline test set set (only the images, without labels) was released on July 2018, and the participants were invited to submit their results for an offline validation. Each participant could receive a maximum of five evaluations on this set. Each task was evaluated separately according to a uniform criteria. In particular:

\subsubsection{Glaucoma classification:} The teams submitted a table with a glaucoma likelihood per each image on the set. A receiver operating characteristics (ROC) curve was created based on the gold standard glaucoma diagnostic, and the area under the curve (AUC) was used as a ranking score for the classification task, $S_\text{class}$ (the higher, the better). 
Additionally, a reference sensitivity $\text{Se} = \frac{\text{TP}}{\text{TP} + \text{FN}}$ value at a specificity $\text{Sp} = \frac{\text{TN}}{\text{TN} + \text{FP}}$ of 0.85 was also reported, with TP, FP, TN and FN standing for true/false positives and true/false negatives, respectively. This value was not taken into account for the ranking, but allowed each team to assess the overall performance of the classification algorithm in a setting when a low number of false positives is tolerated.

\subsubsection{OD/OC segmentation:} The teams submitted one segmentation file for each image. These files were encoded in grayscale BMP format where 0 corresponded to the optic cup, 128 to the optic disc and 255 elsewhere. The results were compared with the gold standard segmentation using the Dice index (DSC) for OD/OC separately, and the mean absolute error (MAE) of the vertical cup-to-disc ratio (vCDR) estimations. In particular, DSC define the overlap between two binary regions:

\begin{equation}
\text{DSC}_k = 2 \frac{|Y_k \cap \hat{Y}_k|}{|Y_k \cup \hat{Y}_k|}
\end{equation}

\noindent where $Y_k$ and $\hat{Y}_k$ are the ground truth and predicted segmentations of the region of interest $k$, respectively (with $k=$ OD or OC). On the other hand, MAE is defined as:

\begin{equation}
\text{MAE} = \text{abs}( \text{vCDR}(\hat{Y}_\text{OC}, \hat{Y}_\text{OD}) - \text{vCDR}(Y_\text{OC}, Y_\text{OD}) )
\end{equation}

\noindent where $\text{vCDR}(\text{OD},\text{OC}) = \frac{d(\text{OC})}{d(\text{OD})} $ is a function that estimates the vCDR based on the vertical diameter $d$ of the segmentations of the OD and the OC, respectively. Each team was ranked using the average value of each of these metrics separately, resulting in three rank values $R^{\text{DSC}_\text{OD}}_\text{segm}$, $R^{\text{DSC}_\text{OC}}_\text{segm}$ and $R^{\text{MAE}}_\text{segm}$,
and an overall segmentation score $S_\text{segm}$ was assigned to each team based on the following weighted average:

\begin{equation}
S_\text{segm} = 0.35 \times R^{\text{DSC}_\text{OD}}_\text{segm} + 0.25 \times R^{\text{DSC}_\text{OC}}_\text{segm} + 0.4 \times R^{\text{MAE}}_\text{segm}.
\end{equation}

\noindent Notice that in this case, a lower $S_\text{segm}$ value is better than a higher one. Since the MAE of the vCDR is calculated based on the segmentation of OC and OD, we set a larger weight for vCDR than to each individual segmentation term. Moreover, it is standard in the literature (Section~\ref{sec:previous-works}) to first segment the OD region and then extract the OC from the cropped OD area. Hence, we assigned a larger weight to the OD segmentation results than to the OC.


An overall offline score was assigned to each method based on:
\begin{equation}
S_\text{val} = 0.4 \times R_\text{class} + 0.6 \times R_\text{segm}
\label{eq:score}
\end{equation}
where $R_\text{class}$ and $R_\text{segm}$ are the team rank positions based on the classification and segmentation scores $S_\text{class}$ and $S_\text{segm}$, respectively. A larger weight was assigned to the ranking for the segmentation task as the vCDR, derived from OD/OC segmentation, can be used as a primary score for glaucoma classification.
An offline test set based leaderboard was created by setting a rank position $R_\text{val}$ for each team, based on $S_\text{val}$. Only those teams that submitted reports describing their proposed approaches were taken into account for this leaderboard. These reports can be easily accessed from the challenge website.~\footnote{\url{https://refuge.grand-challenge.org/Results-Onsite_TestSet/}}

The first 12 teams according to $S_\text{val}$ were invited to attend to the on-site challenge, that was held in person at MICCAI. The test set (only the images) was released during the workshop, and the 12 teams had to submit their results before a time deadline (3 hours). The last submission of each team was taken into account for evaluation. Both an on-site rank and a final rank were assigned to each team. The on-site rank $R_\text{test}$ was created using the scoring described in Eq.~\ref{eq:score}, while the final rank $R_\text{final}$ was based on a score $S_\text{final}$ calculated as the weighted average of the off-line and on-site rank positions:

\begin{equation}
\label{eq:winner}
S_\text{final} = 0.3 \times R_\text{val} + 0.7 \times R_\text{test}.
\end{equation}

\noindent Notice that a higher weight was assigned to the results on the test set. In this paper we only focus on the results obtained on the test set, during the on-site challenge. 

The evaluation was performed using a Python 3.6 open-source framework that was specially developed for the challenge and is publicly available.~\footnote{\url{https://github.com/ignaciorlando/refuge-evaluation}}

%% file: results.tex
\section{Results}
\label{sec:results}



This section presents the results on the REFUGE test set of the 12 teams that participated in the on-site challenge. The official final rankings according to the offline and online test set performances can be accessed on the REFUGE website. 

\subsection{Glaucoma classification}
\label{subsec:results-classification}

\begin{table*}
\centering
\caption{Summary of the glaucoma classification methods evaluated in the on-site challenge, in alphabetical order using the teams names.}
\resizebox{\textwidth}{!}{
\begin{tabular}{C{2.2cm}|C{3cm}|C{4cm}|C{3cm}|C{8cm}|C{2cm}}
  \hline
  \textbf{Team} & \textbf{Inputs} & \textbf{Architectures} & \textbf{Training set} & \textbf{Methodology}  & \textbf{Post-processing} \\
  \hline
  AIML & Full image / ONH area & ResNet-50, -101, -152~\citep{he2016deep}, 38~\citep{wu2019wider} & REFUGE training set & Ensemble of glaucoma likelihoods from multiple networks pre-trained on ImageNet and fine-tuned on REFUGE training set & Ensemble by averaging \\ 
  \hline
  BUCT & ONH area, grayscale & Xception~\citep{chollet2017xception} & REFUGE training set & Training from scratch on grayscale images & None \\
  \hline
  CUHKMED & OD/OC segmentation & None & None & vCDR values computed from ellipses fitted to automated OD/OC segmentations & None \\
  \hline
  Cvblab & Full image & VGG19~\citep{simonyan2014very}, Inception V3~\citep{szegedy2016rethinking}, ResNet-50~\citep{he2016deep}, Xception~\citep{chollet2017xception} & REFUGE training set, DRISHTI-GS, HRF, ORIGA and RIM-ONE r3 & Ensemble of glaucoma likelihoods from multiple networks pre-trained on  ImageNet and fine-tuned, classes in REFUGE training set balanced using SMOTE~\citep{chawla2002smote} & Ensemble by averaging \\
  \hline
  Mammoth & ONH area with CLAHE & ResNet-18~\citep{he2016deep} and CatGAN~\citep{wang2017catgan} & Sample from REFUGE training set & Ensemble of ResNet models pre-trained on ImageNet and fine-tuned using REFUGE data and synthetic images generated with CatGAN & None \\
  \hline
  Masker & Full image & ResNet~\citep{he2016deep} & REFUGE training set and ORIGA & Linear combination of vCDR and predictions of multiple ResNet networks & Ensemble with vCDR \\
  \hline
  NightOwl & ONH area with/without exp. transform & Custom & REFUGE training set (10-fold cross-validation) & Ensemble of classification networks trained to predict glaucoma from features produced by the encoders of the segmentation networks & Ensemble by maximum \\
  \hline
  NKSG & Full image & SENet~\citep{Hu2018} & REFUGE training set (5-fold cross-validation) & SE-Net pretrained on images from Kaggle DR challenge~\citep{kaggledr} and fine-tuned on REFUGE data, best model from cross-validation taken for final prediction & None \\
  \hline
  SDSAIRC & Crop with ONH in upper-left corner & ResNet-50~\citep{he2016deep} & REFUGE training set & Logistic regression classifier trained with vCDR values from OD/OC segmentation and output of ResNet-50 model fine-tuned from ImageNet & None \\
  \hline
  SmileDeepDR & ONH area & DeepLabv3+~\citep{Chen2018} &  REFUGE training set & Adaptation of a segmentation network to predict a glaucoma likelihood & None \\
  \hline
  VRT & Full image with custom mask for attention & Custom~\citep{son2018classification} & Kaggle~\citep{kaggledr}, MESSIDOR~\citep{decenciere2014feedback} and IDRiD~\citep{porwal2018indian} & Attention guided model trained on public data sets of DR images, weakly labelled using pre-trained models for glaucoma classification, RNFL defects detection and segmentation of ONH pathological changes  & None \\
  \hline
  WinterFell & ONH area & ResNet-101, -152~\citep{he2016deep}, DensNet-169, -201~\citep{huang2017densely} & ORIGA & Ensemble of glaucoma likelihoods from multiple networks pre-trained on Image-Net and fine-tuned on ORIGA & Ensemble by mode, max. and min. \\
  \hline
\end{tabular}}
\label{table:submitted-glaucoma-classification} 
\end{table*}

\begin{table}[t]  
\centering
\caption{Classification results of the participating teams in the REFUGE test set. 
The last row corresponds to the results obtained using the ground truth vertical cup-to-disc ratio (vCDR).}
\begin{tabular}{C{1.3cm}|C{3cm}|C{2cm}|C{2cm}}
  \hline
  \textbf{Rank} & \textbf{Team} & \textbf{AUC} & \textbf{Reference sensitivity} \\
  \hline
  \textbf{1} & \textbf{VRT} & \textbf{0.9885} & 0.9752 \\
  2 & SDSAIRC & 0.9817 & \textbf{0.9760} \\
  3 & CUHKMED & 0.9644 & 0.9500 \\
  4 & NKSG & 0.9587 & 0.8917 \\
  5 & Mammoth & 0.9555 & 0.8918 \\
  6 & Masker & 0.9524 & 0.8500 \\
  7 & SMILEDeepDR & 0.9508 & 0.8750 \\
  8 & BUCT & 0.9348 & 0.8500 \\
  9 & WinterFell & 0.9327 & 0.9250 \\
  10 & NightOwl & 0.9101 & 0.9000 \\
  11 & Cvblab & 0.8806 & 0.7318 \\
  12 & AIML & 0.8458 & 0.7250 \\
  \hline
  \multicolumn{2}{c|}{Ground truth vCDR} & 0.9471 & 0.8750 \\
  \hline
\end{tabular}
\label{table:results-detection} 
\end{table}

\begin{figure}[!t]
\centering
\includegraphics[width=0.7\columnwidth]{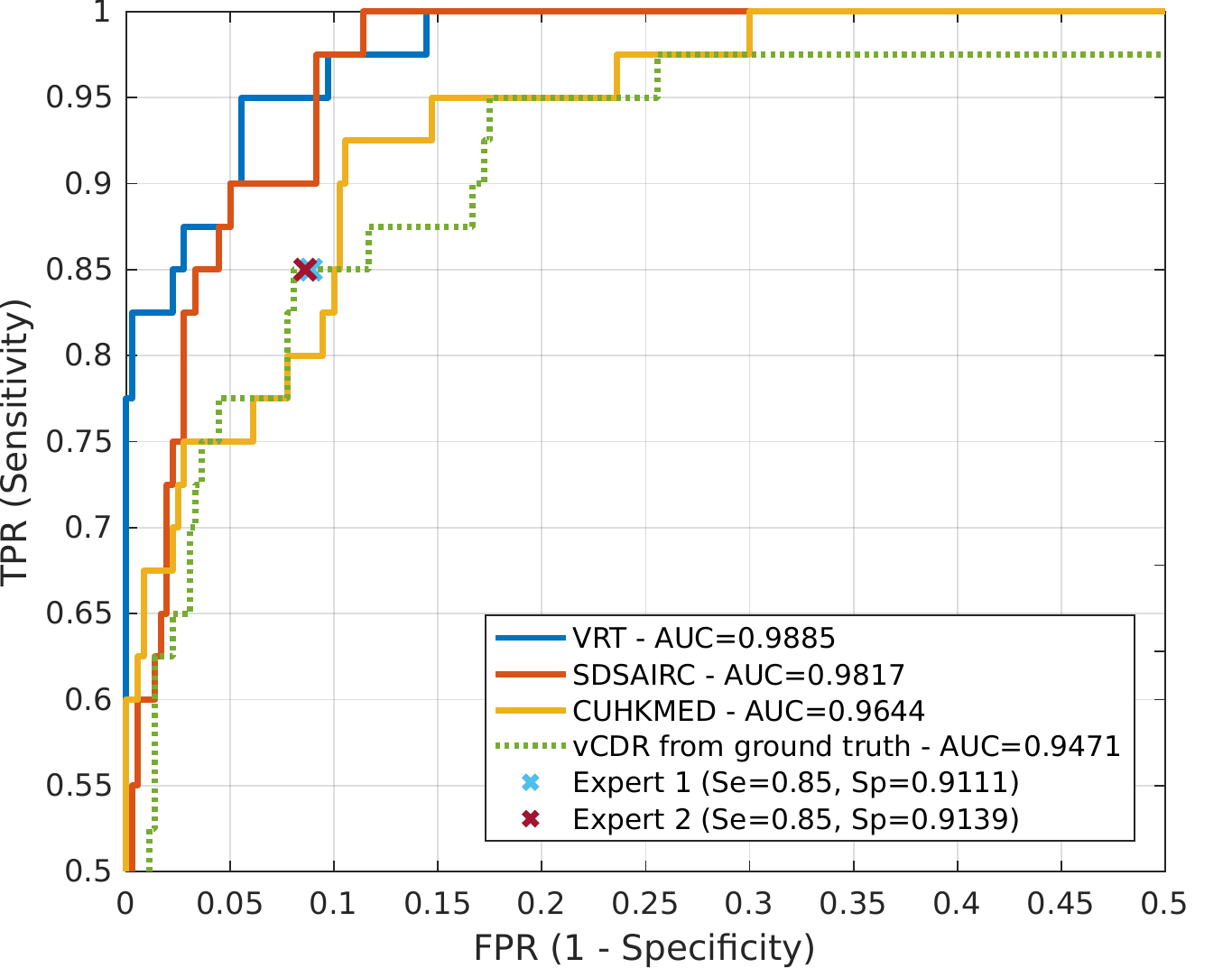}
\caption{ROC curves and AUC values corresponding to the three top-ranked glaucoma classification methods (solid lines) 
and the vertical cup-to-disc ratio (vCDR) (green dotted line). Crosses indicate the operating points of two glaucoma experts.}
\label{fig:roc-curves}
\end{figure}

The participating methods for glaucoma classification are summarized in Table~\ref{table:submitted-glaucoma-classification}. Further details about each method are provided in the appendix. The evaluation of the classification task, in terms of AUC and the reference sensitivity at 85\% specificity, is presented in Table~\ref{table:results-detection}. We also included an additional approach based on using the ground truth vCDR values as a glaucoma likelihood for classification. Figure~\ref{fig:roc-curves} presents the ROC curves of the three top-ranked teams and the ground truth vCDR values. The curves for each participating method are available for downloading in the challenge website. Matt-Whitney U hypothesis tests~\citep{delong1988comparing} with $\alpha = 0.05$ were performed using \cite{vergara2008star} tool, to compare the statistical significance of the differences in the AUC values of these top-ranked teams. VRT reported the best classification performance, achieving significantly better results that the ground truth vCDR ($p = 0.006$). Compared with SDSAIRC and CUHKMED--the second and third teams, respectively--the differences were only significant with respect to CUHKMED (CUHKMED: $p = 0.007$, SDSAIRC: $p = 0.187$). Both SDAIRC and CUHKMED achieved also higher AUC values than the ground truth vCDR, although the differences were not statistically significant ($p > 0.05$). If the results of the best three teams are combined e.g. by normalizing their likelihoods and taking the average as a glaucoma score, the AUC is only marginally improved, with no significant differences with respect to the results of the best team ($p = 0.576$).

In order to understand the relevance of the classification results, a comparison with glaucoma experts was performed. To this end, two independent ophtalmologists visually graded the test set images and assigned a binary glaucomatous/non-glaucomatous label to each of them. These two glaucoma specialists were not part of the group of experts that provided the ground truth labels and did not take part of any discussion regarding data collection/preparation or the organization of the challenge. Notice that no clinical information but only the fundus image was used in this case to perform the annotation. This criteria was followed in order to ensure the same inputs to both the experts and the networks. The sensitivity and specificity values obtained by each human reader are included as expert operating points in Figure~\ref{fig:roc-curves}. The two points are close to each other due to a high level of agreement between the two experts (96.25\% of the cases). The experts graded with the same sensitivity (85\%) and slightly different specificity (91.11\% and 91.39\%) and accuracy (90.50\% and 90.75\%). If only the cases with their consensus are considered, then their joint accuracy increases to 92.21\%, while their joint sensitivity remains the same (85\%) and the specificity reaches 93.04\%.
Despite the fact that both readers agreed with the vCDR curve in terms of sensitivity and specificity, this is pure coincidence as they did not take part of the OD/OC annotation procedure and did not have access to any segmentation.

\begin{figure*}[!t]
\centering
\includegraphics[width=0.98\textwidth]{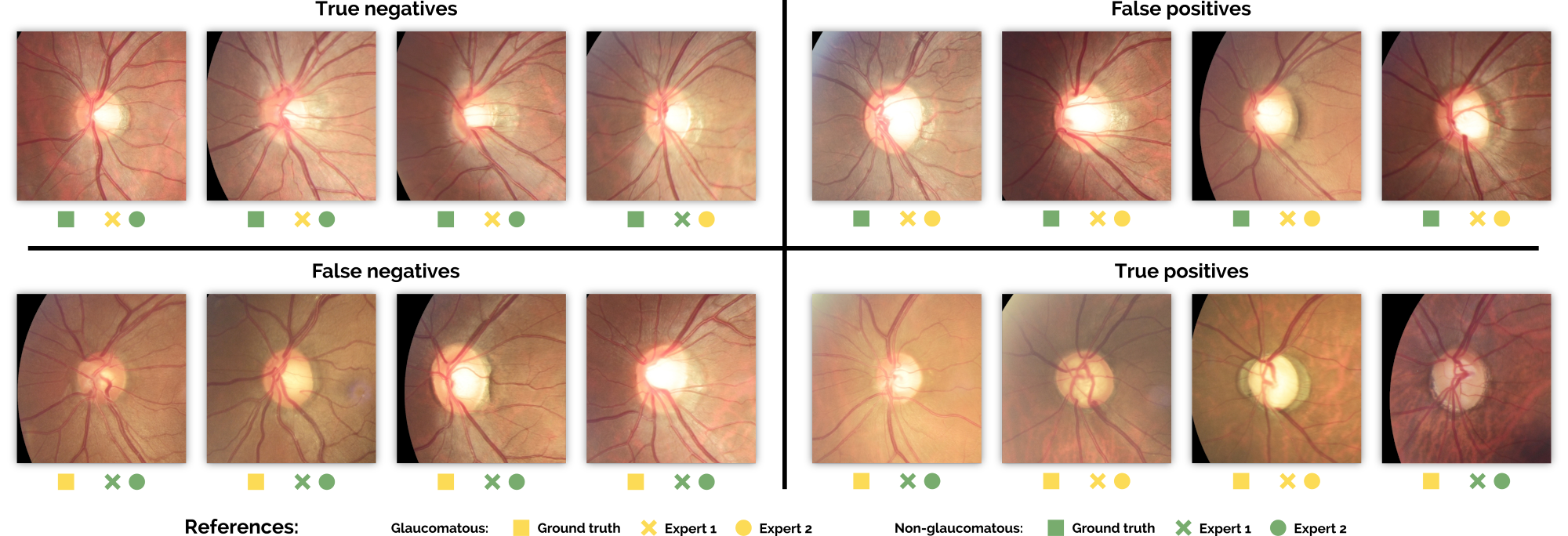}
\caption{Qualitative results for glaucoma classification. Images are zoomed in the ONH area for better visualization. True positives (negatives) correspond to cases in which the ensemble of the three top-ranked methods reported a high (low) score. False positives (negatives) are images for which the ensemble returned a low (high) score. Ground truth and two experts' labels for glaucomatous (yellow) and non-glaucomatous (green) cases are included as colored squares, crosses and circles, respectively.}
\label{fig:qualitative-classification}
\end{figure*}


Figure~\ref{fig:qualitative-classification} illustrates a sample of true negatives, false positives, false negatives and true positive glaucoma detections from the REFUGE test set. The results correspond to the classification performed by the two additional experts and the average of the normalized glaucoma likelihoods of the three top-ranked teams. Since these values are not binary decisions but glaucoma scores, the false positive (negative) images were selected such that their assigned value was higher (lower) when the ground truth label was negative (positive). Similarly, the true positive (negative) images correspond to cases in which the joint likelihood is high (low). 


\subsection{Optic disc/cup segmentation}
\label{subsec:results-segmentation}

\begin{table*}  
\centering
\caption{Summary of the glaucoma classification methods evaluated in the on-site challenge, in alphabetical order using the teams names. FCN(s) stands for fully convolutional network(s).}
\resizebox{\textwidth}{!}{
\begin{tabular}{C{2.2cm}|C{2cm}|C{4cm}|C{3cm}|C{9cm}|C{2.2cm}}
  \hline
  \textbf{Team} & \textbf{Inputs} & \textbf{Architectures} & \textbf{Training set} & \textbf{Methodology}  & \textbf{Post-processing} \\
  \hline
  AIML & Full image  & FCNs ResNet-50, -101, -152~\citep{he2016deep} and -38~\citep{wu2019wider} & REFUGE training set & Two stages: (i) Coarse ONH segmentation with ResNet-50, cropping, (ii) Fine-grain OD/OC segmentation with multi-view ensemble of networks & Ensemble by averaging  \\ 
  \hline
  BUCT & Full image  & U-Net~\citep{ronneberger2015u} & REFUGE training set  & Two stages: (i) OD segmentation with a U-Net, postprocessing, cropping (ii) OC segmentation with U-Net and postprocessing  & OD/OC: largest area element. OD: ellipse fitting.   \\ 
  \hline
  CUHKMED & Full image & U-Net~\citep{ronneberger2015u} and DeepLabv3+~~\citep{Chen2018} & REFUGE training set and validation set (without labels) & U-Net used for cropping, DeepLabv3+ with geometry aware loss and domain shift adaptation via adversarial learning used for final segmentation  & Ensemble by averaging  \\ 
  \hline
  Cvblab & Full image with CLAHE & Modified U-Net~\citep{sevastopolsky2017optic} & DRIONS-DB, DRISHTI-GS, RIM-ONE r3 and REFUGE training set & Two stages: (i) OD segmentation with a modified U-Net, cropping, (ii) OC segmentation with a modified U-Net from cropping & None  \\ 
  \hline
  Mammoth & Full image & Mask-RCNN~\citep{he2017mask} and U-shaped dense network & Sample from REFUGE training set & Two stages: (i) OD segmentation with Mask-RNN and cropping, (ii) OC segmentation with dense U-Net. Resolution restored with spline interpolation & Ensemble of outputs, spline interpolation  \\ 
  \hline
  Masker & Full image &  Mask-RCNN~\citep{he2017mask} & REFUGE training set and ORIGA & Two stages: (i) Mask-RCNN to identify the ONH area, cropping, (ii) Ensemble by bootstrap voting of multiclass Mask-RCNN networks & Ensemble by voting  \\ 
  \hline
  NightOwl & Full image & U-shaped dense network & REFUGE training set  & Two stages: (i) C-Net for ONH detection, matching filter and cropping, (ii) OD/OC segmentation using two F-Nets  & Opening and closing, Gaussian smoothing \\ 
  \hline
  NKSG & ONH area & DeepLabv3+~\citep{Chen2018} & REFUGE training set & Multiclass segmentation using DeepLabv3+ on cropped images pre-processed with pixel quantization & None  \\ 
  \hline
  SDSAIRC & Full image & M-Net~\citep{fu2018joint} & REFUGE training set  & Two stages: (i) OD segmentation with M-Net, cropping, (ii) OC segmentation with M-Net and postprocessing  & Ellipse fitting  \\ 
  \hline
  SmileDeepDR & Full image & U-shaped network with squeeze-and-excitation blocks (X-Unet) & REFUGE training set & X-Unet pre-trained for predicting ground truth labels, and fine-tuned separately for segmenting OD/OC using L1 regression loss & None  \\ 
  \hline
  VRT & Full image & U-Net~\citep{ronneberger2015u} and vessel-based network~\citep{son2017retinal} & IDRiD and RIGA data sets & Two different U-Nets were applied for OD/OC segmentation, respectively. An auxiliary CNN using vessel segmentations as inputs was connected to the U-Nets to aid in the segmentation & Holes filling, convex-hull  \\ 
  \hline
  WinterFell & Full image & Faster R-CNN~\citep{girshick2015fast} and ResU-Net~\citep{shankaranarayana2017joint} & ORIGA & Two stages: (i) ONH detection with Faster R-CNN, (ii) OD/OC segmentation in multiple color spaces with ResU-Net & None \\ 
  \hline
\end{tabular}}
\label{table:submitted-segmentation} 
\end{table*}

\begin{figure*}[!t]
\centering
\subfloat[Optic disc segmentation]{\includegraphics[width=0.49\textwidth]{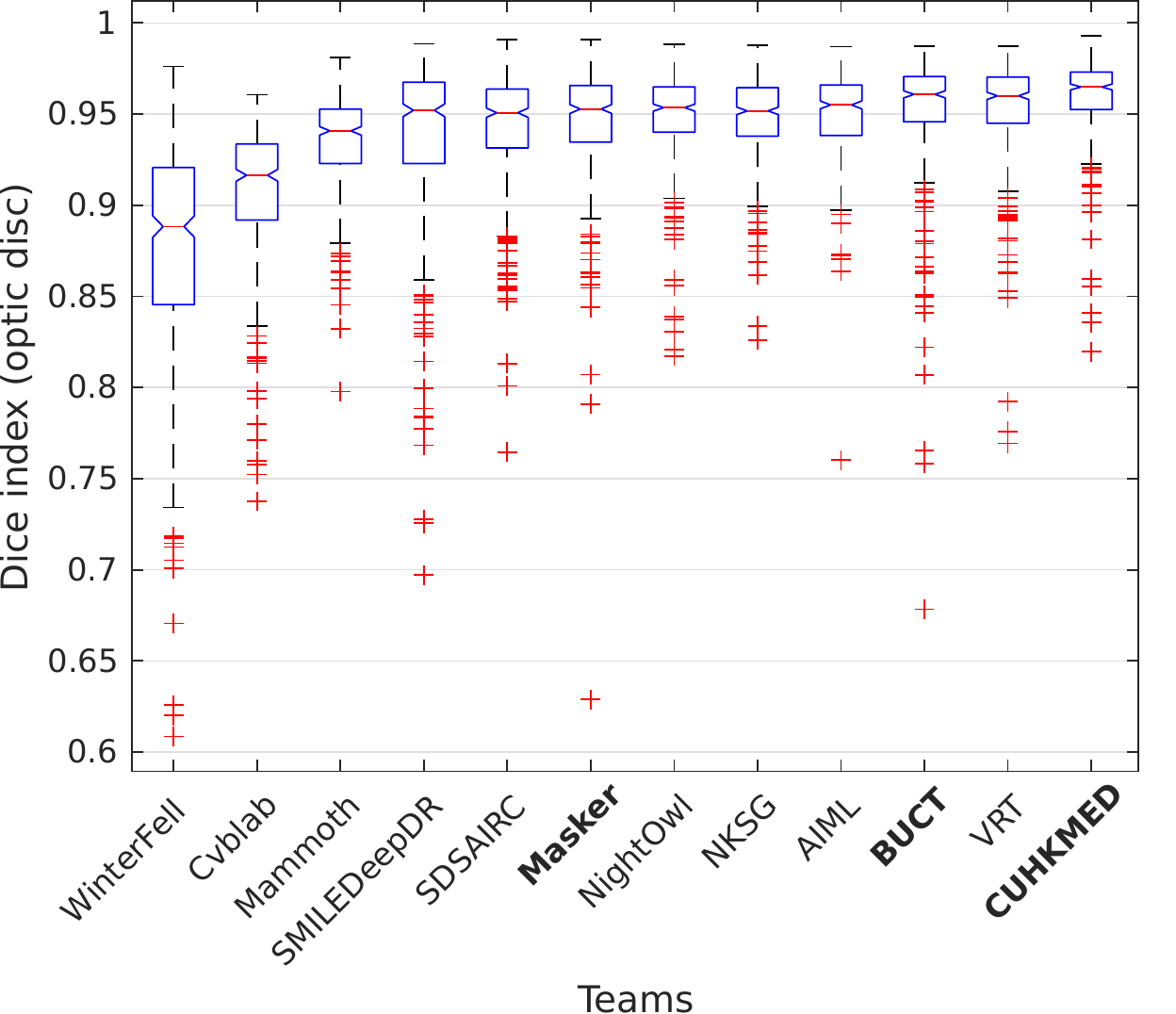}}
\subfloat[Optic cup segmentation]{\includegraphics[width=0.49\textwidth]{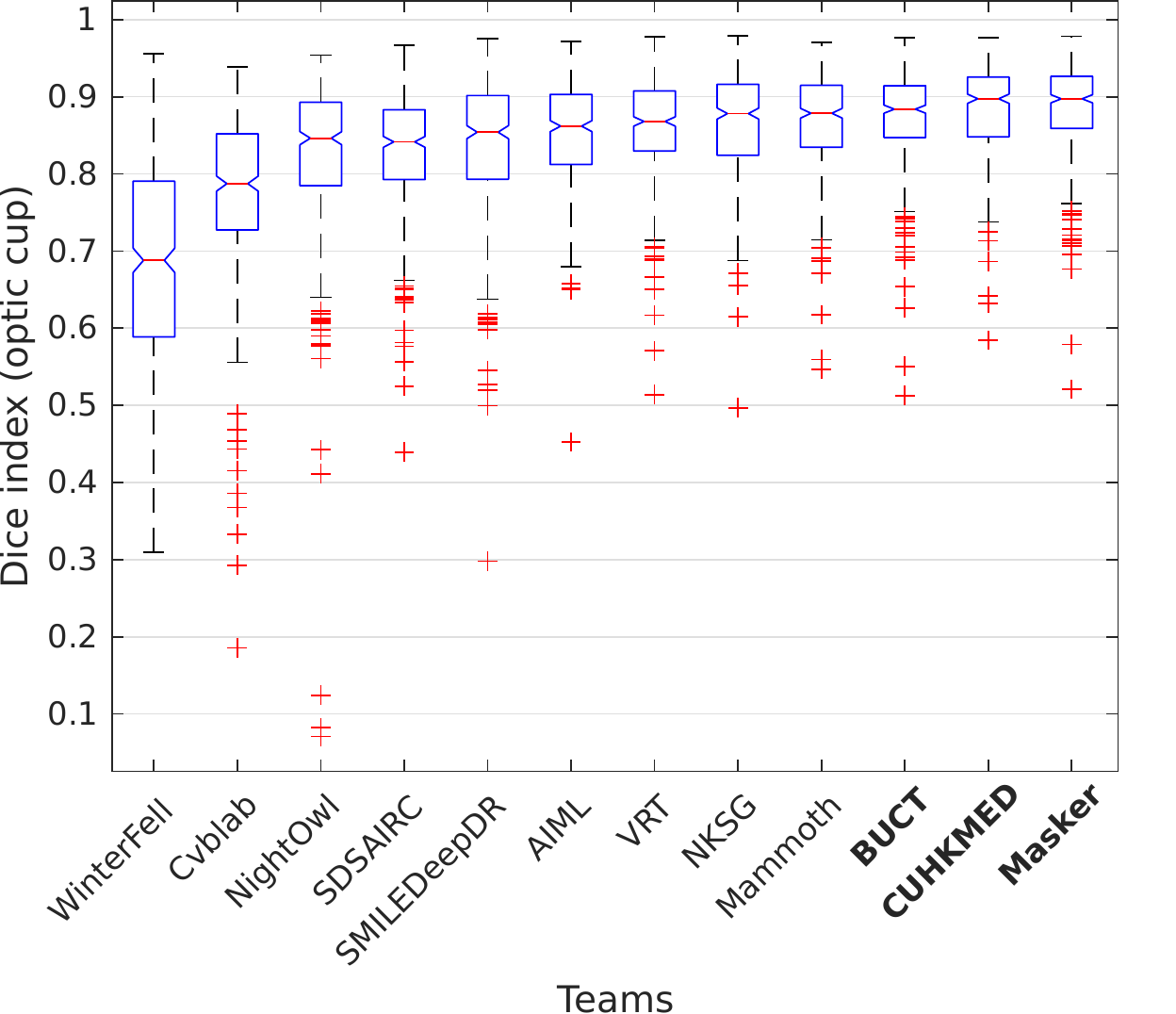}}

\subfloat[MAE for vCDR]{\includegraphics[width=0.49\textwidth]{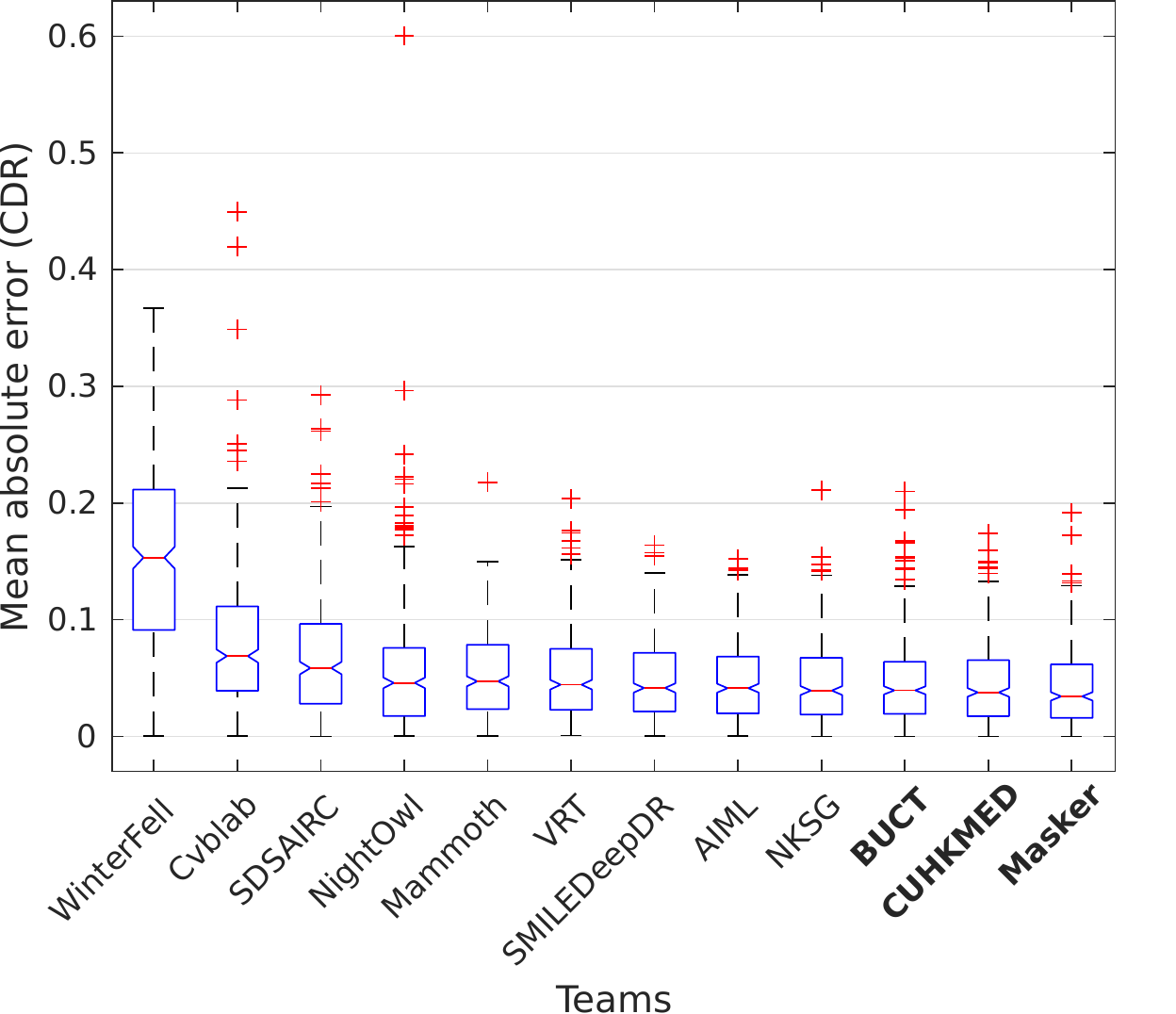}}
\caption{Box-plots illustrating the performance of each optic disc/cup segmentation method in the REFUGE test set. Distribution of Dice (DSC) values for (a) optic disc and (2) optic cup, and (c) mean absolute error (MAE) of the estimated vertical cup-to-disc-ratio (vCDR). The three top-ranked teams in the final leaderboard (CUHKMED, Masker and BUCT) are highlighted in bolds.} 
\label{fig:box-plots-segmentations}
\end{figure*}

\begin{table}  
\centering
\caption{Optic disc/cup segmentation results in the REFUGE test set. Average Dice (Avg. DSC) index for optic cup and disc and mean absolute error (MAE) of the vertical cup-to-disc ratio (vCDR). Teams are sorted by their final rank.} 
\begin{tabular}{C{1cm}|c|c|C{1.2cm}|C{1.2cm}|C{1.2cm}|C{1.2cm}|C{1.2cm}|C{1.2cm}}
  \hline
  \multirow{2}{*}{\textbf{Rank}} & \multirow{2}{*}{\textbf{Team}} & \multirow{2}{*}{\textbf{Score}} & \multicolumn{2}{c|}{\textbf{Optic cup}} & \multicolumn{2}{c|}{\textbf{Optic disc}} & \multicolumn{2}{c}{\textbf{vCDR}} \\
  \cline{4-9}
   & & & \textbf{Rank} & \textbf{Avg. DSC} & \textbf{Rank} & \textbf{Avg. DSC} & \textbf{Rank} & \textbf{MAE} \\
  \hline
   \textbf{1} & \textbf{CUHKMED}     & \textbf{1.75} & 2  & 0.8826 & \textbf{1}  & \textbf{0.9602} & 2  & 0.0450 \\
   2 & Masker      & 2.5  & \textbf{1}  & \textbf{0.8837} & 7  & 0.9464 & \textbf{1}  & \textbf{0.0414} \\
   3 & BUCT        & 3    & 3  & 0.8728 & 3  & 0.9525 & 3  & 0.0456 \\
   4 & NKSG        & 4.6  & 5  & 0.8643 & 5  & 0.9488 & 4  & 0.0465 \\
   5 & VRT 	   	   & 5.4  & 6  & 0.8600 & 2  & 0.9532 & 7  & 0.0525 \\
   6 & AIML    	   & 5.45 & 7  & 0.8519 & 4  & 0.9505 & 5  & 0.0469 \\
   7 & Mammoth 	   & 7.1  & 4  & 0.8667 & 10 & 0.9361 & 8  & 0.0526 \\
   8 & SMILEDeepDR & 7.45 & 4  & 0.8367 & 10 & 0.9386 & 8  & 0.0488 \\
   9 & NightOwl    & 8.6  & 10 & 0.8257 & 6  & 0.9487 & 9  & 0.0563 \\
   10 & SDSAIRC    & 9.15 & 9  & 0.8315 & 8  & 0.9436 & 10 & 0.0674 \\
   11 & Cvblab     & 11   & 11 & 0.7728 & 11 & 0.9077 & 11 & 0.0798 \\
   12 & WinterFell & 12   & 12 & 0.6861 & 12 & 0.8772 & 12 & 0.1536 \\
   \hline
\end{tabular}
\label{table:results-segmentation} 
\end{table}

The evaluated methods for OD/OC segmentation are briefly described in Table~\ref{table:submitted-segmentation}. The interested reader could refer to the appendix for further details. The distribution of DSC and MAE values obtained by each of the participating teams in the REFUGE test set are represented as boxplots in Figure~\ref{fig:box-plots-segmentations}. Table~\ref{table:results-segmentation} summarizes the final ranking, based on the average performance of each team. The statistical significance of the differences in performance of the top-ranked teams was assessed by means of Wilcoxon signed-rank tests ($\alpha = 0.05$). CUHKMED reported the highest DSC values for OD segmentation, significantly outperforming all the alternative approaches ($p < 1.41 \times 10^{-7}$). VRT and BUCT achieved the second and third higher average DSC values, respectively. However, their performance was not statistical significantly different with respect to each other ($p = 0.734$). For OC segmentation, Masker reported the highest average DSC value, followed by CUHKMED and BUCT. The differences in the DSC values achieved by Masker were statistically significant with respect to every other team ($p < 1 \times 10^{-4}$), except to CUHKMED ($p = 0.387$). When evaluating in terms of MAE of the vCDR estimation, Masker also reported the best results, consistently outperforming every other method ($p < 0.014$). CUHKMED retained the second place, although with no significant differences with respect to the BUCT ($p < 0.403$), which was ranked in the third place.


To study the complementarity of the three top-ranked methods according to the final leaderboard (CUHKMED, Masker and BUCT), a majority voting segmentation was obtained from their results, both for OD and OC. By quantitatively evaluating the resulting segmentations, and comparing to the constitutive models, we observed significant improvement in DSC values for OC (mean $\pm$ std = 0.8922 $\pm$ 0.0551, Wilcoxon signed rank test, $p < 1.91 \times 10^{-7}$) and OD (mean $\pm$ std = 0.9626 $\pm$ 0.0196, Wilcoxon signed rank test, $p < 1.07 \times 10^{-7}$). When the estimated vCDR values were analyzed in terms of MAE (mean$\pm$std = 0.0398$\pm$0.0313), the improvements were statistically significant compared to CUHKMED and BUCT ($p < 1.27 \times 10^{-4}$) but not to Masker ($p = 0.148$).

\begin{figure}[!t]
\centering
\includegraphics[width=0.7\textwidth]{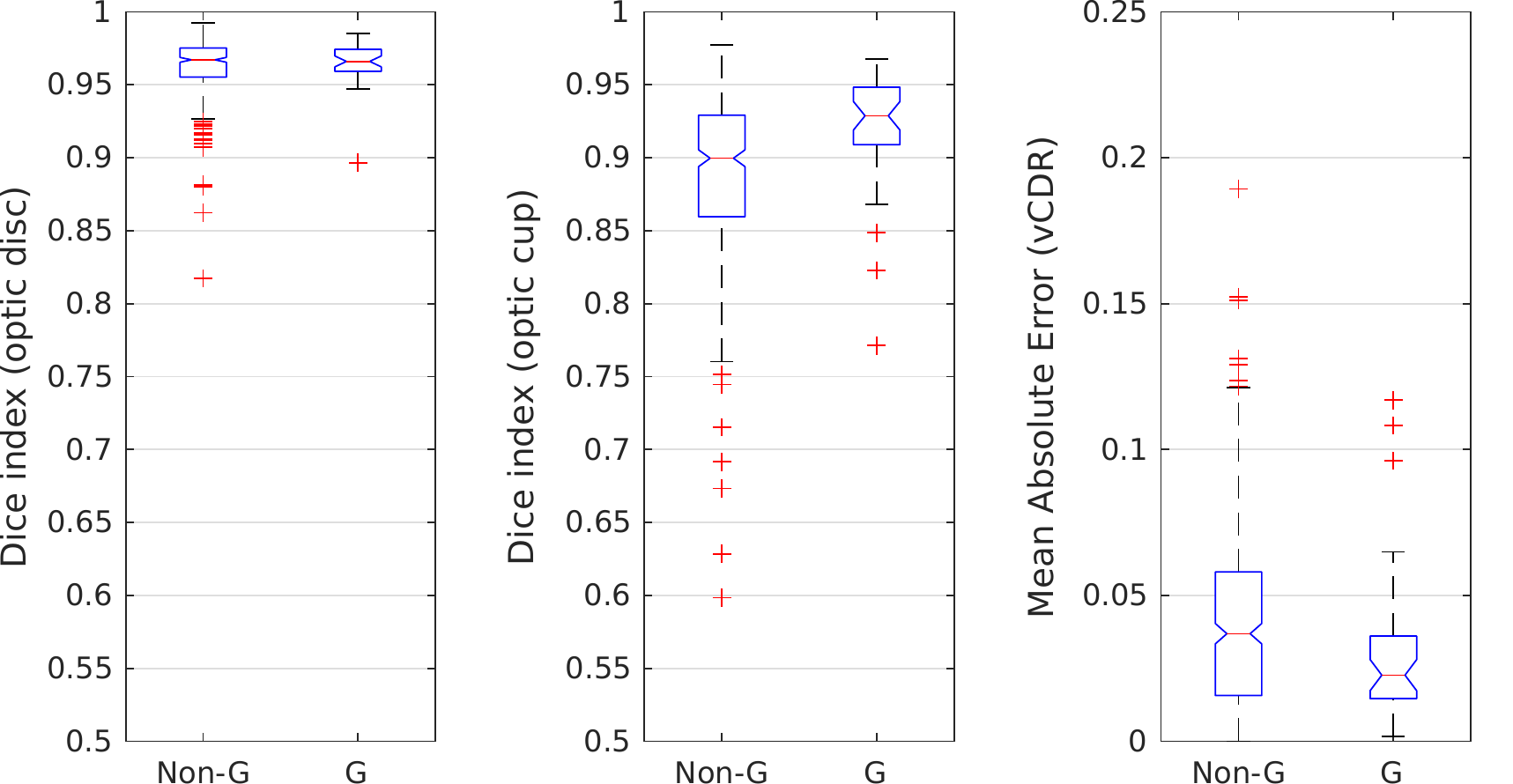}
\caption{Segmentation metrics stratified for the glaucomatous (G) and non-glaucomatous (Non-G) subsets in the REFUGE test set. From left to right: Dice values for optic disc and optic cup segmentation, and mean absolute error of vertical cup-to-disc ratio (vCDR) estimates. The performance values were computed from segmentations obtained by majority voting of the top-three methods (CUHKMED, Masker and BUCT).}
\label{fig:stratified-segmentation}
\end{figure}


Figure~\ref{fig:stratified-segmentation} presents the distribution of DSC and MAE values stratified according to the glaucomatous/non-glaucomatous ground truth labels of the images. These metrics were calculated from the majority voting segmentations obtained from the three winning teams (CUHKMED, Masker and BUCT), although an analogous behavior was observed when stratifying the individual results of the methods. The statistical significance of the differences between groups was assessed using a Wilcoxon rank-sum test due to the unpaired nature of the two sets (360 vs. 40 samples, respectively). For OD segmentation, the differences in performance between the two groups were not statistically significant ($p = 0.3435$). Higher values were obtained for OC segmentation in the glaucomatous group ($p = 2.09 \times 10^{-5}$), while the MAE values were significantly smaller in the positive set ($p = 0.023$).

\begin{figure*}[!t]
\centering
\includegraphics[width=0.98\textwidth]{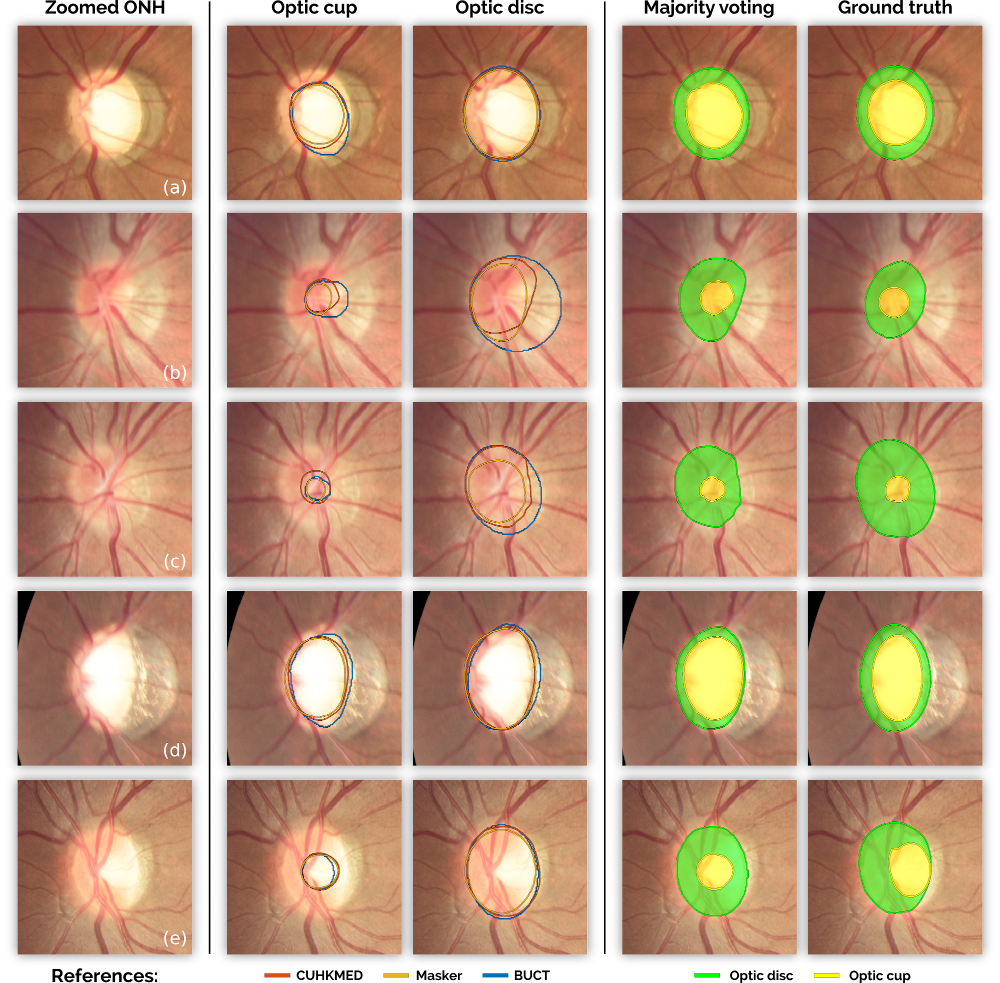}
\caption{Optic disc/cup segmentation results in the REFUGE test set. From left to right: zoomed ONH area, segmentation results from the three top-ranked teams (BUCT, Masker and CUHKMED) for optic cup and disc segmentation, majority voting of these methods and ground truth segmentations.}
\label{fig:segmentation-qualitative}
\end{figure*}

Finally, Figure~\ref{fig:segmentation-qualitative} presents some qualitative examples of the segmentations of the top-three ranked methods and those obtained by majority voting: (a) and (d) present some degree of peripapillary atrophy (PPA), (b) and (c) correspond to cases with ambiguous edges and (c) and (e) are the worst performing cases as measured in terms of DSC for the OD and the OC, respectively. The general behavior of each of the methods is rather stable compared with each other for most of the cases (Figure~\ref{fig:segmentation-qualitative} (a), (d) and (e)). In challenging scenarios such as those observed in Figure~\ref{fig:segmentation-qualitative} (b-e), where the edges of the ONH structures are difficult to assess, majority voting between methods resulted in more accurate segmentations. However, the voting only made a significant difference when the methods were complementary (Figure~\ref{fig:segmentation-qualitative} (b) and (c) vs. (d) and (e)).

%% file: discussion.tex
\section{Discussion}
\label{sec:discussion}


The key methodological findings concluded after analyzing the challenge results are discussed in Section~\ref{subsec:findings}. Subsequently, Section~\ref{subsec:strengths-and-limitations} covers challenge strengths and limitations that might be taken into account in future editions. Finally, Section~\ref{subsec:clinical-implications} covers the clinical implications of the results.

\subsection{Findings}
\label{subsec:findings}

Our unified evaluation framework allowed us to draw some technical findings that might be useful for future developments in the field. Section~\ref{subsec:discussion-classification-methods} and Section~\ref{subsec:discussion-segmentation-methods} describe our findings in the classification and segmentation tasks, respectively. A special analysis of ensemble methods is provided in Section~\ref{subsec:discussion-ensemble}.

\subsubsection{Classification methods}
\label{subsec:discussion-classification-methods}
In line with the evolution of the literature in the field, we observed that the proposed solutions for glaucoma detection were generally based on state-of-the-art convolutional neural networks for image classification, with the only exception of SMILEDeepDR and CUHKMED (Table~\ref{table:submitted-glaucoma-classification}). SMILEDeepDR adapted a segmentation network to predict both the OD/OC regions and a glaucoma likelihood, based on the intermediate feature representation generated by the architecture. CUHKMED, on the other hand, proposed to use a normalized vCDR predicted from the OD/OC segmentations. 

The classification networks comprised of general-purpose image classification models that were top-ranked in ImageNet Large Scale Visual Recognition Competition~\citep{russakovsky2015imagenet}, such as VGG19~\citep{simonyan2014very}, ResNets~\citep{he2016deep}, DenseNets~\citep{huang2017densely}, Inception V3~\citep{szegedy2016rethinking} or Xception~\citep{chollet2017xception}, among others.
Since training such deep architectures from scratch on a training set with only 400 images might be prone to overfitting, most of the teams initialized the CNNs with pre-trained weights from ImageNet and fine-tuned them afterwards using the CFPs. Alternatively, NKSG team used pre-trained weights from the Kaggle DR data set~\citep{kaggledr}. This eases the fine-tuning task as the transition from natural images to fundus photographs is less smooth than the one from images of DR to glaucoma. Only BUCT trained its networks from scratch, although using the ONH area and not the full images. Nevertheless, we observed that the best solutions were based not only on the application of an existing classification network but also using domain-specific heuristics as discussed next. 

CUHKMED achieved the third place by relying only on the prediction of the vCDR. Deep learning was in this case used indirectly, as it was applied for segmenting the OD/OC areas. Exploiting a well-known, clinical parameter such as the vCDR allowed them to identify most of the cases with cupping, which usually correspond to advanced glaucomatous damage. SDSAIRC (second place), on the other hand, obtained better results by combining vCDR estimates with glaucoma likelihoods provided by different CNNs. Team Masker (sixth place) followed a similar idea but using a network trained on full images. Instead, SDSAIRC trained the CNNs using a cropped version of each image in which the ONH is observed at the upper-left corner. We hypothesize that this configuration indirectly forces the network to identify other complementary signs that are not captured by the vCDR, such as the presence of peripapillary hemorrhages--which appear in the border of the OD (Figure~\ref{fig:glaucoma-detection-challenging} (b))--or RNFL defects--observed as striated patterns spanning from the ONH (Figure~\ref{fig:glaucoma-detection-challenging} (c)). 
Similarly, the winning team, VRT, further improved this idea by using an attention-guided network~\citep{son2018classification}. This approach takes as input both a fundus image and a region mask covering the optic disc and the RNFL area. By means of a structural region separation model~\citep{park2018novel}, the network is driven to analyze regions in which disease-specific biomarkers may occur. In principle, a classification network with enough capacity would learn to identify abnormal image patterns by itself, without needing an attention mask, although this is highly dependent on the size of the training set~\citep{poplin2018prediction}. VRT team instead restricted the field-of-view of the method by focusing on disease-relevant areas. This attention mechanism might help to learn more accurate classification models that does not require manual annotations of glaucoma-related abnormalities such as RNFL defects or peripapillary hemorrhages. On the other hand, VRT increased REFUGE training set by incorporating images from other public data sets, assigning to them image-level classification labels using a pre-trained model. Using additional public data with weak labels was accepted by the organizers as the resulting increased data set has annotations that are still prone to errors. Hence, it was possible to evaluate the contribution of a weak training signal to the proposed approach. The results of VRT seems to empirically show that increasing the training set with further scans is beneficial even if the training labels were obtained automatically.

\subsubsection{Segmentation methods}
\label{subsec:discussion-segmentation-methods}
The proposed solutions for OD/OC segmentation were all based on at least one fully convolutional neural network (Table~\ref{table:submitted-segmentation}). U-shaped networks inspired by the U-Net~\citep{ronneberger2015u} were the prevalent solutions, although incorporating recent technologies such as residual connections (AIML), atrous convolutions (BUCT) or multiscale feeding inputs (SDSAIRC), among others. Most of the strategies were also based on the two stage approach described in Section~\ref{sec:previous-works} of first roughly identifying the ONH and then performing the OD/OC segmentation on a cropped version of the original image. 
The three top-ranked teams followed this principle. CUHKMED and BUCT used a classical U-Net~\citep{ronneberger2015u} to localize the ONH area, while Masker applied a Mask-RCNN~\citep{he2017mask}. Once this area was localized, CUHKMED segmented the OD/OC using a DeepLabv3+~\citep{Chen2018} architecture, a recently published approach based on atrous separable convolutions that is able to capture multiscale characteristics. Masker, on the other hand, used an ensemble of Mask-RNNs trained with bootstrap, while BUCT used a classical U-Net. NKSG was ranked fourth using the same architecture as CUHKMED, but normalizing image appeareance between training and offline test sets using a pixel quantization technique. CUHKMED, on the other hand, accounted for this domain shift using adversarial learning, which could explain its better performance.

Interestingly, we noticed that the three top-ranked methods and their ensemble by majority voting achieved consistently better segmentation results in the subset of glaucomatous subjects than in the non-glaucomatous cases. This can be linked with the fact that advanced glaucoma cases with severe cupping usually present more clear interfaces between the OD and the OC. Such a characteristic would explain why the improvement is more evident in the Dice index obtained for the OC than in the performance for OD segmentation. On the other hand, the segmentation models showed a slightly worst accuracy in challenging scenarios with unclear transitions between the OD/OC, such as those illustrated in Figure~\ref{fig:segmentation-qualitative} (b), (c) and (e). The lack of depth information in monocular color fundus photographs turns this task significantly difficult. Research in developing automated methods for predicting depth maps from CFPs is currently ungoing, trying to correlate image features with ground truth labels obtained from other imaging modalities such as stereo fundus photography~\citep{shankaranarayana2019fully} or OCT~\citep{thurtell2018local}. These techniques might aid to solve ambiguities in these scenarios.

If the segmentation results are analyzed separately, BUCT and CUHKMED achieved the second and the third place for OC segmentation and the first and third places for OD segmentation, respectively (Table~\ref{table:results-segmentation}). Using the same criteria, Masker achieved the first place for OC segmentation but the seventh for OD segmentation. Surprinsingly, the team reported the lowest MAE of the vCDR estimation. This indicates that most of their errors in the OD prediction occurs horizontally, and therefore not affect the prediction of its vertical diameter. 

\subsubsection{Ensemble methods}
\label{subsec:discussion-ensemble}

Independently of the target task, we noticed that several submissions exploited to some extent the application of ensembles. Combining the outcomes of multiple models is a common practice in challenges as it allows to achieve (sometimes marginal) quantitative improvements that can eventually ensure higher positions in the final rankings~\citep{kaggledr,kamnitsas2017ensembles}. We observed three types of ensembles in REFUGE. Teams AIML, Cvblab and WinterFell, for instance, combined the outputs of multiple architectures trained with the same data set. This approach allows to take advantage of the characteristics of each model without explicitly integrating them into a single network. Alternatively, team Mammoth averaged the outputs of a single architecture trained under different configurations (e.g. images preprocessed with multiple strategies). Under this setting, model selection is bypassed as there is no need to choose a single configuration because a subset or even all of them are exploited in test time. Finally, a similar approach was followed by NightOwl and Masker for classification and segmentation, respectively, although by training the same architecture on different portions of the training data. 


Applying majority voting or averaging on the collective responses of multiple models might ensure more reliable results. This has been recently applied by \cite{de2018clinically} for retinal OCT analysis with a significant success. However, this will strictly depend on the complementarity (and non-redundance) of the ensembled methods. 
We experimentally assessed how complementary the top-winning methods are by averaging their normalized likelihoods (for the glaucoma classification task) and taking segmentations by majority voting (for the OD/OC segmentation task). In both tasks we have observed increments in performance that in principle indicate that each winning approach is complementary with the others. This was more notorious for the second task (Figure~\ref{fig:box-plots-segmentations}), where the segmentations obtained by majority voting of the top-ranked methods were more accurate when the models disagreed the most. This indicates that, despite their impressive but similar performance, the methods are still complementary with each other, and can be integrated to generate a more accurate automated response. This can be qualitatively observed in the segmentation examples in Figure~\ref{fig:segmentation-qualitative}, where e.g. BUCT oversegmented the OD and the OC in (b) but achieved more accurate results in (c). On the other hand, cases such as those in Figure~\ref{fig:segmentation-qualitative} (d) and (e) illustrate the need of model diversity to achieve more accurate results under challenging conditions. The improvements in the classification task were only marginal when averaging the top-three models (AUC = 0.9901) and not significant ($p = 0.576$). This is most likely a consequence of the high agreement between the models, indicating that there are still cases that are missclassified. In any case, notice, however, that we cannot argue that the ensemble of these particular approaches is \textit{per-se} the best way to go for performing the individual tasks. To ensure a proper generalization error and avoid any selection bias, an ensemble approach must be based on models that are choosen according to their individual performance on a held-out validation set.

\subsection{Challenge strengths and limitations}
\label{subsec:strengths-and-limitations}

REFUGE was the first open initiative aiming to introduce a uniform evaluation framework to assess automated methods for OD/OC segmentation and glaucoma classification from CFPs. To this end, the challenge provided to the community with the largest public available data set of fundus photographs (1200 scans) to date. In addition, it contains gold standard clinical diagnostic labels, and a high quality reference OD/OC masks and fovea positions from a total of nine glaucoma experts. This unique characteristic ensures a more appropriate development of glaucoma classification methods, as it was recently observed that training with fundus-derived labels have a negative impact on performance to detect truly diseased cases~\citep{phene2018deep}. To the best of our knowledge, the most similar data set to REFUGE was ORIGA~\citep{zhang2010origa}, which provided 650 images with OD/OC segmentations and glaucoma labels. However, at the time of submitting this manuscript ORIGA was not available anymore\footnote{\url{http://imed.nimte.ac.cn/origa-650.html}}, while, more than 350 teams have successfully registered to the REFUGE website to access the database, with 183 requests submitted after the on-site challenge. Such a large interest of the scientific community in accessing REFUGE data clearly demonstrates that a quality open glaucoma data set and challenge was needed.

The challenge design matched most of the principles for evaluating retinal image analysis algorithms proposed by \cite{trucco2013validating}. In particular, REFUGE data set can be easily accessed through a website that is  part of the Grand Challenges organization. Furthermore, an automated tool is provided to evaluate the results of any participating team, ensuring a uniform, un-biased criterion for comparing methods, based on trustable and accurate annotations. Furthermore, the data is already partitioned into fixed training, offline and online test sets, with labels publicly available only for the first two sets. Future participants are invited to submit their results to the website to estimate their performance on the test set. By keeping these ground truth annotations private we prevent the teams to overfit on test data, ensuring a fair comparison between models. 

As the offline test set set was used to determine which teams were qualified to participate in the on-site challenge, the access to the validation labels was initially restricted. Only five submissions per team were allowed to evaluate the performance on the offline test set, limiting its applicability for design tasks such as model selection. As a consequence of this constrain, most of the teams ended up using the REFUGE training set or other third-party data sets for this purpose, which might have affected their performance. To overcome this issue, we have publicly released the offline test set labels right after the onsite event. We encourage future participants to use this data as a validation set, not only for model selection but also to better explain their models' behavior e.g. through ablation studies, to empirically show the contribution of each decision in intermediate results. This might help to better identify good practices to follow when designing glaucoma classification and OD/OC segmentation methods.

Another remark regarding the data set organization is that the winners of the challenge were selected according to a weighted sum of their rankings in the offline and the onsite test sets (Eq.~\ref{eq:winner}). This was intentionally done to reward the participants for their efforts in having good results in the offline test set, while preventing dummy submissions with the sole purpose of participating in the onsite event. This last point was also guaranteed by inviting the 12 best performing teams on the offsite test set to participate in the onsite challenge. Each team was allowed to request a maximum number of 5 evaluations in the offline test set, to avoid strong overfitting on it. Nevertheless, and despite the fact that a low weight was assigned to this rank, the final score might be biased due to some form of overfitting on the offline test set. This paper is focused only on the results of the onsite test set, though, which was held out during the entire challenge and for which only a single submission is allowed. Our conclusions remain therefore unbiased by this issue. Future challenges might perhaps consider the posibility of using four splits instead of three: two with public labels for training and model selection/validation, and other two with private labels for offline and onsite evaluations. Hence, if only one submission is allowed for the last two sets, then further conclusions regarding the generalization ability of the methods could be drawn.

One limitation of REFUGE is the lack of diverse ethnicities in its data set, as the images correspond to a Chinese population. Ethnicities manifest differently in CFP due to changes in the pigment of the fundus. Therefore, it cannot be ensured that the best performing models on the REFUGE challenge can be applied to a different population and obtain the same outcomes without retraining. Furthermore, it is worth mentioning that the percentage of glaucoma cases in the REFUGE data set is higher than expected to be encountered in a screening scenario and more representative of a clinical one. Furthermore, despite the fact that REFUGE data set is the largest publicly available image source for glaucoma classification, 1200 images is still not big enough for developing general enough deep learning solutions. Similar initiatives in other diseases have provided larger data sets: the Kaggle challenge in diabetic retinopathy grading, for instance, released more than 80.000 CFPs for training and testing the algorithms \citep{kaggledr}. The high quality of the images also hampers the applicability of the proposed methods in real screening scenarios, where imaging artifacts and low quality scans are expected to appear much more frequently. A representative screening data set should include comorbidities, diverse ethnicities, ages and genders and low quality images with acquisition artifacts. These characteristics should be addressed in future challenges e.g. by multicenter collaboration for data collection, to ensure that the winning models can be applied in a more general environment.

Another potential limitation of REFUGE data sets is that the OD/OC manual segmentations were performed from CFPs, without considering depth information or knowledge about the Bruch’s membrane opening. The latter is considered by most as the best OD anatomical delimitation, and serves as reference for one of the most recent measures of the amount of retinal nerve fibers, the Bruch’s membrane opening minimum rim width (BMO-MRW)~\citep{reis2012optic}. As a consequence, these annotations might be deviated from the real anatomy of both areas. In an effort to alleviate this drawback, our annotations resulted from majority voting of delineations performed first by seven different glaucoma specialists, and then controlled by another independent expert. We believe that these steps ensured much more reliable outcomes than using annotations from a single reader, although further validation would be certainly needed to confirm this hypothesis. Better ground truth labels could be obtained e.g. by delineating OD/OC from OCT scans, which provide cross-sectional images of the retina (and therefore depth information). However, the resulting labels should be afterwards transferred to CFP e.g. via image registration, which might be subject to errors if the registration algorithm fails. In that case, manual correction based on CFP would be still required, and deviation from the true geometry might then still occur.

REFUGE data was prepared with glaucoma status as the main target label. After applying the pre-defined protocol to analyze the follow-up medical records of each CFP, the images were annonimized and it is unfeasible now to link them with their clinical information. As a consequence, additional labels for other co-existing morbities in the non-glaucomatous and glaucomatous sets were lost. Similar initiatives might take this into consideration in the future, and provide not only the target labels for the specific applications of the challenge but also complementary information such as labels for other conditions or functional parameters such as the IOP. This would not only allow further assessment of the challenge results (e.g. the influence of comorbidities or some parameters in the final outcomes) but also indirectly benefit other derived applications (e.g. automated myopia or megalopapilae detection or computerized prediction of IOP from CFP).

Finally, it is important to remark a point regarding the evaluated proposals and their differences in training settings, particularly those related with data availability. Despite the fact that ORIGA is claimed to be publicly available since 2010~\citep{zhang2010origa}, by the time of this publication it was not possible to download the images. These kind of fluctuations in data access might have influenced the decisions made by the participating teams about which image sources to use for training their models. Future challenges might address this issue e.g. by providing a curated list of potential sources to retrieve images. In any case, it is worth mentioning that only one of the top-ranked teams (Masker, who achieved the second place in the onsite evaluation of the segmentation task) used ORIGA to train its model, so in principle the access to this data was not per se a guaranty of success.

\subsection{Clinical implications of the results and future work}
\label{subsec:clinical-implications}

Can we envision automated systems for detecting suspicious cases of glaucoma from fundus photographs? This is still an open question, although REFUGE results might help us to catch a glimpse of a possible answer. With the constant development of much cheaper and easy-to-use fundus cameras, it is expected that this imaging technique will be widespread even more in the decades to follow. Turning it into a cost-effective imaging modality for glaucoma screening is still pending due to the subtle manifestation of the early stages of the disease in these images. Nevertheless, novel image analysis techniques based on deep learning can pave the way towards computer-aided screening of glaucoma from fundus photographs. 

We observed that some of the proposed segmentation models were able to obtain accurate vertical cup-to-disc ratio estimates. The best team in the segmentation task (CUHKMED) achieved the third place in the classification ranking by using the vCDR as a glaucoma likelihood, with sensitivity and specificity values almost in pair with two human experts, and statistically equivalent to those obtained using the ground truth measurements. The best performing teams, however, complemented ONH measurements with the classification outcomes of deep learning based models, and were able to significantly surpass the glaucoma experts, with increments in sensitivity up to a 10\%. Although these results are limited to a specific image population, we can still argue that these deep learning models are able to identify complementary features, invisible to the naked eye, that are essential to ensure a more accurate diagnosis of the disease. Representing the activation areas on the images might help to better understand which areas were considered by the automated models to produce their predictions. We believe that these tools might contribute in the future to a better identification of glaucoma suspects based on color fundus images alone. 

The challenge results also seem to indicate that vCDR, although being an important risk factor for glaucoma, is not enough for detecting the disease at a single time-point basis. This is likely as a consequence of vCDR ignoring other important features such as ONH hemorrhages or RNFL defects. Other metrics derived from the OD/OC relative shapes were recently observed to outperform vCDR for screening, such as the rim to disc ratio  \citep{kumar2019rim}. Notice also that some clinical guidelines such as \cite{guidelines} do not recommend vCDR to classify patients, as several healthy discs might have large vCDR. Attention is instead recommended towards the neuroretinal rim thickness and the degree of vCDR symmetry between eyes~\citep{guidelines}. In any case, vCDR is still a relevant parameter (it achieved an AUC of 0.9471 in our test set for glaucoma classification) that might help to analyze disease progression (e.g. in a follow-up study in which the evolution of the vCDR is assessed for each visit of the patient). Glaucoma screening tools should certainly not ignore vCDR but should also take other biomarkers into account such as the presence, size and location of ONH hemorrhages or the presence and size of RNFL defects, to ensure more reliable predictions.

The complementarity of CFP and OCT for automated glaucoma screening still needs to be exploited. Although CFP allows a cost-effective assessment of the retina, features such as the damage in the ONH or the RNFL are more evident in optic disc centered OCT. This is due to the fact that OCT provides a three dimensional view of the retina, with a micrometric resolution. Hence, the cross-sectional scans--or B-scans--can be used to quantify the thickness of the RNFL or the degree of cupping in the ONH. Nevertheless, the OCT acquisition devices are more expensive than fundus cameras, and the manual analysis of the volumetric information is costly and time-consuming. Developing deep learning methods to quantify glaucoma biomarkers from OCT scans is therefore necessary to complement results in fundus images and pave the way towards cost-effective glaucoma screening.

%% file: conclusions.tex
\section{Conclusions}

We summarized the results and findings from REFUGE, the first open challenge focused on glaucoma classification and optic disc/cup segmentation from color fundus photographs. We analyzed the performance of each of the twelve teams that participated in the on-site edition of the competition, during MICCAI 2018. We observed that the best approaches for glaucoma classification integrated deep learning techniques with well-known glaucoma specific biomarkers such as changes in the vertical cup-to-disc ratio or retinal nerve fiber layer defects. The two top-ranked teams, on the other hand, achieved better results than two glaucoma specialists, a promising sign towards using automated methods to identify glaucoma suspects with fundus imaging. For the segmentation task, the best solutions took into account the domain shift between training and test sets, aiming to regularize the models to deal with image variability. Cases with ambiguous edges between the optic disc and the optic cup showed to be the most challenging ones. Further research should be performed to improve the results in those scenarios. For both tasks of the challenge, we observed that integrating the outcomes of multiple models allowed to improve their individual performance. 

REFUGE unified evaluation framework allowed us to identify good common practices based on the results of the twelve proposed approaches. We expect these findings to help in the future to develop strong baselines for comparison and to aid in the design of new automated tools for image-based glaucoma assessment.

REFUGE challenge data and evaluation framework are publicly accessible through the Grand Challenges website at \url{https://refuge.grand-challenge.org/}. In parallel, a sibling platform has been deployed at \url{http://eye.baidu.com/} with capabilities to automatically process teams' submissions. Future participants are invited to submit their results in any of these websites. Participation requests have to include all the requested information (full real name, institution and e-mail) to be approved, or will be otherwise declined. The two websites will remain permanently available for submissions, to encourage future developments in the field.